\documentclass[11pt,a4paper]{article}

\usepackage[utf8]{inputenc}
\usepackage[T1]{fontenc}
\usepackage[margin=1in]{geometry}
\usepackage{hyperref}
\usepackage{booktabs}
\usepackage{amsfonts}
\usepackage{amsmath}
\usepackage{amssymb}
\usepackage{microtype}
\usepackage{xcolor}
\usepackage{graphicx}
\usepackage{multirow}
\usepackage{cleveref}
\usepackage{float}
\usepackage[section]{placeins}
\usepackage{needspace}
\usepackage{caption}
\usepackage{tcolorbox}
\tcbuselibrary{breakable,skins}
\usepackage{textcomp}
\usepackage{newunicodechar}
\usepackage{natbib}
\usepackage{listings}
\usepackage{newtxtext,newtxmath}
\usepackage{enumitem}
\usepackage{cleveref}
\usepackage{titlesec}

\titleformat{\section}
  {\large\bfseries}
  {\thesection.}{0.6em}{}

\titleformat{\subsection}
  {\normalsize\bfseries}
  {\thesubsection.}{0.6em}{}

\titleformat{\subsubsection}
  {\normalsize\itshape}
  {\thesubsubsection.}{0.6em}{}

\titlespacing*{\section}{0pt}{1.2ex plus 0.4ex minus 0.2ex}{0.6ex}
\titlespacing*{\subsection}{0pt}{1.0ex plus 0.3ex minus 0.2ex}{0.4ex}

\lstdefinestyle{promptbox}{%
  basicstyle=\footnotesize\ttfamily,
  breaklines=true,
  breakatwhitespace=false,
  frame=single,
  rulecolor=\color{gray!50},
  backgroundcolor=\color{gray!8},
  xleftmargin=0.5em,
  xrightmargin=0.5em,
  aboveskip=4pt,
  belowskip=4pt,
  keepspaces=true,
  showstringspaces=false,
  columns=flexible,
}

\newunicodechar{−}{\textminus}

\newtcolorbox{agentbox}[2][]{
  colback=#1!6, colframe=#1!60, fonttitle=\bfseries\small,
  title={#2}, breakable, left=4pt, right=4pt, top=3pt, bottom=3pt,
  boxrule=0.5pt
}
\definecolor{buyercolor}{RGB}{25,118,210}
\definecolor{sellercolor}{RGB}{211,47,47}

\graphicspath{
  {./}
  {../experiments/results/qwen3_235b/exp_asym/plots/}
  {../experiments/results/qwen3_235b/exp_asym/think_accuracy/plots/}
  {../experiments/results/qwen3_235b/exp_trade_plan/plots/}
  {../experiments/results/qwen3_235b/exp_trade_plan/think_accuracy/plots/}
  {../experiments/results/qwen3_235b/exp_counterfactual/plots/}
}

\newcommand{\utilde}{\tilde{U}}
\newcommand{\cond}[1]{\texttt{#1}}

\title{\vspace*{-1.5cm} Counterparty Modeling is Not Strategy: \\ The Limits of LLM Negotiators}
\author{
Romain Cosentino \;\;\; Sarath Shekkizhar \;\;\; Adam Earle \;\;\; Silvio Savarese \\
\vspace{0.2em}
\small Salesforce AI Research \\
}
\date{}

\begin{document}
\maketitle

\begin{abstract}
Negotiation requires more than inferring what the other side wants: it requires using that information to make advantageous offers and counteroffers over multiple turns. We study whether large language model (LLM) agents do this in a controlled multi-attribute bargaining environment. We find that current LLM agents can model a counterparty’s preferences, but do not reliably turn that knowledge into strategic bargaining. When given negotiating partner preference information, agents model it accurately and early in their reasoning traces, yet this does not reliably improve outcomes for the informed side. Turn-level analyses show why: agents often respond to what they believe the counterparty values, but do not consistently pair those moves with gains on their own high-value attributes. Sellers are more accommodating overall, and in asymmetric-information conditions, the informed side often makes the more weakly compensated concessions. Because agents fail to leverage this underlying utility structure for strategic advantage, their final agreements are heavily dictated by surface-level opening anchors rather than actual utility weights. Finally, requiring agents to explicitly state concession-for-reciprocity trades before making an offer makes individual turns look more strategic, but ultimately fails to improve the efficiency of the final agreements.
\end{abstract}
\vspace{.5cm}
\section{Introduction}
\label{sec:intro}

Negotiation is a compact test of strategic social intelligence \citep{zhou2023sotopia}. To negotiate well, an agent must not only infer what the other party wants, but also decide how to use that information over multiple turns to maximize their own outcome: which concessions to make, which attributes to hold fixed, and what to demand in return. For example, knowing the counterparty's priorities should allow an agent to concede on low-cost attributes in exchange for gains on its own top priorities. Strategic use of preference information therefore means more than identifying what the other party wants: it means turning that knowledge into reciprocal exchanges that improve one’s own position.

This makes negotiation a useful setting for separating two capabilities that are often conflated in large language models (LLMs): \emph{counterparty modeling} and \emph{strategic utilization} \citep{zhao2024antgptlargelanguagemodels, Kosinski_2024}. Information about the other party's preferences should help only if it can be converted into better contingent action. The central question of this paper is whether current LLM agents can make that transition from counterparty modeling to strategy.

As LLMs are increasingly used in procurement, contracting, sales, and organizational coordination \citep{song2026large, kirshner2026talking}, it is not enough for an agent to recognize another party's preferences. It must convert that knowledge into advantageous action under interaction. An agent that understands the other party but systematically bargains poorly may appear socially intelligent while remaining strategically fragile and exploitable.

This setting highlights a general issue in LLM evaluation: evidence that a model can describe or reason about a task-relevant variable should not be taken as evidence that it can also use that variable effectively in sequential behavior \citep{huang2025sibenchbenchmarkingsocialintelligence}. In humans, that inference is often reasonable. In LLM agents, however, these two capacities may be only partially aligned \citep{junque2025llms}. Negotiation offers a controlled setting in which to examine that distinction directly.

Recent work has established that LLMs can participate in negotiations, respond to social tactics, and sometimes achieve strong final outcomes in bilateral bargaining or benchmark settings \citep{deng2024bargaining, bianchi2024negotiationarena, fu2023improving, hua2024gametheoretic, liu2026agenticpay, zhu2025automated, rana2024whenai, lewis2017deal, davidson2024evaluating}. But these studies mostly evaluate \emph{what} agreements are reached (final prices, deal rates, welfare, or broad measures of efficiency and fairness) rather than whether an agent can systematically leverage advantages, such as asymmetric information, to secure superior individual outcomes. Because the focus has remained largely on these final states, much less is known about \emph{how} LLM agents use information across turns. In particular, prior work does not isolate whether LLM agents convert explicit knowledge of their counterparty's preferences into reciprocal, strategically effective bargaining behavior over the course of a negotiation.

In this paper, we study whether current LLM agents can use  information about the other side’s preferences strategically in multi-attribute bilateral negotiation. Specifically, we place agents in a
controlled car purchase bargaining environment with ten contract attributes and
randomized linear utility profiles. Each agent's utility is a private weighted sum over
contract features, determining both their preferences across outcomes and their
reservation value. Because this is a familiar consumer-negotiation scenario, it helps reduce domain comprehension confounds and lets us focus on strategic reasoning. Across conditions, we vary whether the buyer, the seller, both, or neither receives a prompt-level ranked summary of the other party’s preferences. We then analyze
behavior at three levels: final outcomes, reasoning trace beliefs, and turn-level
bargaining dynamics.

Our main finding is not simply that information about a negotiating partner’s preferences has limited effects on outcomes, but rather that even when modeled correctly, it does not reliably translate into a strategic bargaining advantage. While providing information modestly improves joint welfare, it does not reliably benefit the information holder. In particular, providing the seller with the buyer's preferences does not help the seller extract value, and instead, the seller often simply accommodates the buyer.

The reasoning traces show that this is not a failure of perception. Informed agents frequently identify the negotiating partner’s priorities early and explicitly. The problem arises in the transition from modeling to strategy. To diagnose that transition, we introduce metrics that measure whether offers move in the direction the agent believes the other party prefers, and whether those moves are paired with gains on the agent's own high-value attributes. Our results show that the agents respond to what the other party values, but do not reliably turn that information into reciprocal exchange.

Because agents fail to leverage this underlying utility structure for strategic advantage, their negotiations are strongly influenced by the first concrete values introduced in the conversation. Consistent with this pattern, final agreements are much more strongly tied to the first proposed price than to the agents' actual utility weights. This suggests that models rely too much on specific numbers in the prompt or dialogue, instead of using the latent utility structure that should guide efficient bargaining.

We then investigated whether this gap could be closed by forcing agents to explicitly structure their trades. We introduced a trade-plan intervention that requires agents to use a give/ask template before making an offer. However, this forced structure likewise does not improve efficiency. This suggests that the missing ingredient is not merely the ability to articulate a one-step trade, but the ability to embed those trades in reciprocal multi-turn bargaining.

These results point to a broader conclusion: current LLM agents can model negotiation-relevant information, but do not reliably use it to organize bargaining strategically across turns. Even when an agent has accurate information about the negotiating partner’s priorities, that information does not consistently translate into reciprocal exchange that improves the agent’s own position.
 \vspace{-.25cm}
\paragraph{We summarize our contributions as follows:}
\begin{itemize}[nosep]
    \item Access to the negotiating partner’s preferences does not reliably translate into bargaining advantage for the informed side: outcome effects are modest and often benefit the other side instead (Sec.~\ref{sec:exp_asym}).
    \item This failure is not explained by weak counterparty  modeling. Informed agents form early beliefs about the other party's priorities, indicating that the breakdown lies in how those beliefs are translated into bargaining behavior (Sec.~\ref{sec:think_accuracy}).
    \item Turn-level analyses reveal that agents act on what they believe the negotiating partner values, but do not reliably convert those moves into gains on their own priorities (Sec.~\ref{sec:gap}).
    \item Requiring explicit give/ask trade plans via a template does not make negotiation more effective, suggesting that the missing capability is not the ability to state a trade, but to embed such trades in reciprocal multi-turn bargaining (Sec.~\ref{sec:trade_plan}).
\end{itemize}

\section{Related Work}
\label{sec:related}

\paragraph{LLMs as negotiators.}
Several recent papers show that LLMs can negotiate non-trivially in bilateral settings. This line of work builds on earlier negotiation-dialogue work such as \citet{lewis2017deal}, which showed that end-to-end dialogue agents can learn to negotiate in natural language and that rollout-based planning can improve performance. \citet{deng2024bargaining} study buyer-seller bargaining under light prompting and show that LLMs can achieve high trade rates, negotiate prices close to theory-guided benchmarks, and exploit asymmetric information in scalar-price bargaining. \citet{bianchi2024negotiationarena} introduce \emph{negotiationarena}, a platform covering ultimatum games, trading games, and price negotiations, and show that tactics such as aggression can change final payoffs. \citet{fu2023improving} study self-play and AI feedback in a single-item buyer-seller game, showing that some models can improve final deal prices across rounds. \citet{zhu2025automated} analyze fully automated agent-to-agent negotiations in consumer markets and identify large model-dependent performance gaps and multiple failure modes, including budget violations and unreasonable deals. \citet{liu2026agenticpay} broaden the scope further, introducing a benchmark spanning bilateral to many-to-many markets and showing persistent limitations in long-horizon negotiation. These papers establish that LLM negotiation is both feasible and behaviorally rich. They also complement more recent work framing negotiation as a useful testbed for language-model agency \cite{davidson2024evaluating}. 

\paragraph{Game-theoretic evaluation and negotiation workflows.}
A related line of work studies LLMs in game-theoretic environments. \citet{hua2024gametheoretic} evaluate LLM rationality across complete- and incomplete-information games and propose structured workflows that improve performance in tasks such as Deal-or-No-Deal. Relatedly, \citet{lore2024strategic} study the strategic behavior of large language models in canonical two-player social-dilemma games, while \citet{gandhi2023strategic} show that language models can be prompted to perform strategic reasoning in games involving other agents, hidden information, and competing objectives. These results show that LLMs can display nontrivial strategic behavior and that explicit game-theoretic scaffolds can improve rationality and near-optimal allocations.



\paragraph{Strategic reasoning with language models.}
Beyond negotiation specifically, a growing literature studies whether LLMs can reason strategically in settings involving other agents, hidden information, and competing objectives. \cite{gandhi2023strategic} show that pretrained language models can be prompted to perform strategic reasoning in games and can exhibit human-like negotiation strategies in realistic scenarios. 
\cite{liao2024efficacy} further show that self-play can substantially improve language-model performance in negotiation-style non-zero-sum settings. 


\paragraph{Automated negotiation and bargaining theory.}

Additionally, our work sits within a broader academic literature surrounding automated negotiation, bilateral trade, and bargaining under incomplete information \citep{myerson1983efficient, chatterjee1983bargaining, faratin1998negotiation, he2018dealornodeal, baarslag2013evaluating}. Classical models provide clean benchmarks for information, efficiency, and equilibrium, but abstract away from natural-language reasoning and the richer issue structure of modern LLM interactions. 
\\
\\
While the prior work discussed above establishes that LLMs can negotiate, follow workflows, and model game-theoretic scenarios, these studies primarily evaluate final outcomes, e.g., deal rates, prices, overall rationality, or broad measures of efficiency. Even when negotiations unfold over multiple turns, the main object of study is typically the eventual agreement. Recent work has begun to examine the structural mechanisms driving these outcomes, for instance, \citet{wang2026discoveringdifferencesstrategicbehavior} used automated program discovery to show that frontier LLMs maintain highly sophisticated opponent models in simple sequential games like iterated Rock-Paper-Scissors. Our paper takes a different approach by isolating a specific, process-level phenomenon in complex, natural-language bargaining: whether agents that can infer a counterparty's preferences actually convert that knowledge into reciprocal, strategically coupled bargaining behavior over time. Rather than asking if a workflow improves final payoffs, or if models can exploit simple transition matrices, we analyze belief traces, turn-level offer changes, and the coupling between concessions and self-gain. In doing so, we demonstrate that in multi-attribute language-mediated negotiation, the ability to model a partner and the ability to execute a multi-turn strategy remain two distinct and unaligned capabilities.

\section{Experimental Design}
\label{sec:setup}

We study controlled bilateral negotiation in a multi-attribute car-purchase domain. The goal of the design is to isolate strategic information use: agents face the same action space and bargaining protocol, while information about negotiating partner preferences is manipulated experimentally.

\subsection{Multi-Attribute Negotiation Environment}

We use a multi-attribute car-purchase domain with 10 negotiation terms (Table~\ref{tab:terms}).
The feature vector $\phi(x) \in \mathbb{R}^{20}$ is built in two steps.
First, each continuous term is rescaled to $[0,1]$ within the \emph{agent's own feasible range} (e.g., for the buyer on a Sedan, $\texttt{price\_norm} = (\text{price} - 20)/(45 - 20)$, where $20$ and $45$ are respectively the lowest and highest permitted price for the buyer).
Second, categorical and binary terms are one-hot encoded as binary indicators. After both steps every component of $\phi(x)$ lies in $[0,1]$.

\begin{table}[h]
  \centering
  \small
    \captionsetup{justification=centering}
  \caption{
    Negotiation terms and feasible ranges.}
  \label{tab:terms}
  \begin{tabular}{llcc}
    \toprule
    Term & Type & Buyer range & Seller range \\
    \midrule
    Price (\$k)  & continuous & 20--45 & 25--55 \\
    Delivery (days)        & continuous & 1--30  & 7--60  \\
    Down payment (\%)      & continuous & 0--30  & 15--40 \\
    Trade-in (\$k)         & continuous & 5--15  & 0--10  \\
    \midrule
    Model      & categorical & \multicolumn{2}{c}{Sedan, SUV, Truck} \\
    Color      & categorical & \multicolumn{2}{c}{White, Black, Silver, Blue, Red} \\
    Interior   & categorical & \multicolumn{2}{c}{Standard, Premium, Luxury} \\
    Warranty   & categorical & \multicolumn{2}{c}{none, basic, extended} \\
    Service    & categorical & \multicolumn{2}{c}{none, annual, comprehensive} \\
    Accessories & binary     & \multicolumn{2}{c}{true / false} \\
    \bottomrule
  \end{tabular}
\end{table}

Each agent has a private utility function $U_i(x) = \theta_i^\top \phi(x)$. The weights $\theta_i$ are drawn per feature group from uniform distributions with role-specific sign constraints, e.g., the price weight is always negative for the buyer and positive for the seller. For categorical variables, one option is sampled as preferred and assigned a positive weight, while the remaining options receive fractional negative weights. This encoding makes categorical choices contrastive: the preferred option increases utility, whereas the alternatives reduce it relative to that preference. All weights are then L1-normalised, $\|\theta_i\|_1 = 1$, also note that $U_i$ is \emph{not} bounded below by zero: negative weights (e.g.\ the buyer's price weight) contribute negatively when the corresponding feature is large, so the worst-case contract typically gives $U_i < 0$.
Each agent also has a reservation value $d_i$ computed as the utility at the worst feasible contract, and we constrain that a deal requires $U_i(x) > d_i$ for both parties to be concluded.
In the experiments, we report normalized utility
$
\utilde_i = \frac{U_i - d_i}{\max_x U_i(x) - d_i} \,\in\, [0,\,1],$
so $\utilde_i = 0$ is the walk-away point and $\utilde_i = 1$ is the agent's ideal outcome.

\subsection{Models and Traces}
All experiments use Qwen3-235B-A22B \citep{qwen3} and are replicated with DeepSeek-R1-671B \citep{deepseek} in Appendix~\ref{app:deepseek}. For each conversation turn, we log the resulting \texttt{<think>} traces together with the templated (JSON) deal proposition. These traces allow us to analyze the intermediate beliefs and plans that accompany the offers. This is central to our setting, because we ask whether partner knowledge is merely modeled or strategically deployed in multi-turn negotiation. 

\subsection{Information Conditions and Prompt Interventions}
We run two experiment families, each designed to isolate a different part of our central question: whether LLM agents merely \emph{model} partner information, or actually \emph{use} them strategically in negotiation (Table~\ref{tab:conditions}).

\paragraph{Asymmetric information (\cond{exp\_asym}).}
Our first experiment asks the basic question of value-of-information in bargaining: when one side knows more about the other's preferences, does that knowledge improve its outcome? We vary who receives partner preference information derived from the partner's utility weights (neither agent, only the buyer, only the seller, or both) and run 100 trials per condition with randomized utility weights. This experiment tests whether negotiating partner information helps the agent that receives it, and whether any resulting gains reflect strategic use of that information rather than a general increase in agreement or coordination. When an agent receives the other side preference information, it appears in the prompt as a ranked preference list (critical, important, flexible, see Appendix~\ref{app:prompts:prefs} for prompts). Thus, informed agents observe a coarse ranked summary of the counterparty’s higher-priority attributes rather than the full underlying utility weights or utility function.

\paragraph{Trade-plan intervention (\cond{exp\_trade\_plan}).}
Our second experiment asks whether the observed gap can be reduced by imposing explicit reciprocal trade structure in the prompt. We introduce a reasoning template that requires agents, before each offer, to specify a feature to concede, a feature to demand in return, and a concrete give/ask template (Appendix~\ref{app:prompts:tradeplan}). We then compare negotiation with and without this intervention under uninformed and fully informed conditions. This experiment tests whether enforcing a simple concession for reciprocity template is sufficient to convert partner knowledge into more strategically effective multi-turn bargaining.

\begin{table}[tb]
  \centering
    \captionsetup{justification=centering}
  \caption{%
    Experimental conditions investigated.
  }
  \label{tab:conditions}
  \small
  \begin{tabular}{llccr}
    \toprule
    Experiment & Condition & Buyer & Seller  & Trade plan \\
    \midrule
    \multirow{4}{*}{\cond{exp\_asym}}
      & \cond{symmetric\_none}  & \texttimes & \texttimes & \texttimes \\
      & \cond{buyer\_informed}  & \checkmark & \texttimes & \texttimes \\
      & \cond{seller\_informed} & \texttimes & \checkmark & \texttimes \\
      & \cond{symmetric\_full}  & \checkmark & \checkmark & \texttimes \\
    \midrule
    \multirow{4}{*}{\cond{exp\_trade\_plan}}
      & \cond{uninformed\_no\_plan}   & \texttimes & \texttimes & \texttimes \\
      & \cond{uninformed\_with\_plan} & \texttimes & \texttimes & \checkmark \\
      & \cond{informed\_no\_plan}     & \checkmark & \checkmark & \texttimes \\
      & \cond{informed\_with\_plan}   & \checkmark & \checkmark & \checkmark \\
    \bottomrule
  \end{tabular}
\end{table}

\subsection{Metrics}
\label{sec:metric}
To assess both negotiation dynamics and final outcomes, we use the following evaluation metrics.
\paragraph{Outcome metrics.}
We report deal rate, normalized utilities $\utilde_b, \utilde_s$ (buyer and seller respectively), social welfare $\utilde_b + \utilde_s$, distance to the Nash Bargaining Solution $d_\text{NBS}$, and distance to the nearest Pareto-efficient contract $d_\text{Pareto}$.

\paragraph{Belief accuracy.}

For each agent turn, we use an LLM to read the \texttt{<think>} block and extract structured beliefs about the counterparty:
$\{(\text{feature}, \text{direction})\}$.
We compute \emph{signed-accuracy@k}: a prediction counts as correct when it identifies a feature in the partner's true top-k and the correct preference direction.

\paragraph{Strategic coupling.}
Outcome metrics cannot distinguish strategic use of partner knowledge from generic accommodation.
We therefore construct per-turn measures that jointly characterize whether the agent acts on its negotiating partner model \emph{and} extracts value for itself.

\emph{Concession toward counterparty} $c_{T+1}$ measures how much the agent acted on what it believes the other party wants.
For each feature $f$ that the agent's \texttt{<think>} block mentions when reasoning about the negotiating partner, $d_f \in \{-1,+1\}$ is the extracted belief direction (e.g.\ ``the partner wants lower price'' $\Rightarrow$ $d_f = -1$ for price) and $\Delta_f$ is the change in the agent's own offered value from turn $T$ to $T{+}1$.
The product $d_f \cdot \Delta_f$ is positive precisely when the offer moved in the direction the agent believes the counterparty prefers. Summing positive contributions over all mentioned features gives:
\begin{equation}
  c_{T+1} = \textstyle\sum_{f \in \mathcal{M}_{T+1}} \max(0,\, d_f \cdot \Delta_f)
  \label{eq:concession}
\end{equation}
where $\mathcal{M}_{T+1}$ is the set of partner features mentioned in the \texttt{<think>} block at turn $T{+}1$.

\emph{Own gain} $g_{T+1}$ measures, independently of the partner, whether the agent simultaneously captured value on its \emph{own} top priorities.
Let $\theta_{i,1},\ldots,\theta_{i,K}$ be the agent's $K$ largest utility weights (by magnitude) and $\Delta_k$ the change in the offered value on those features.
Then $\operatorname{sign}(\theta_{i,k})\cdot\Delta_k > 0$ whenever the offer moved in the agent's own preferred direction on feature $k$:
\begin{equation}
  g_{T+1} = \textstyle\frac{1}{K}\sum_{k=1}^{K} \operatorname{sign}(\theta_{i,k})\cdot\Delta_k
  \label{eq:owngain}
\end{equation}

A strategically rational agent links the two: it concedes on features the partner values ($c > 0$) while simultaneously extracting gains on its own priorities ($g > 0$). 

\section{Counterparty Information Does Not Reliably Benefit the Informed Side}
\label{sec:exp_asym}

\begin{figure}[h]
  \centering
  \includegraphics[width=0.7\linewidth]{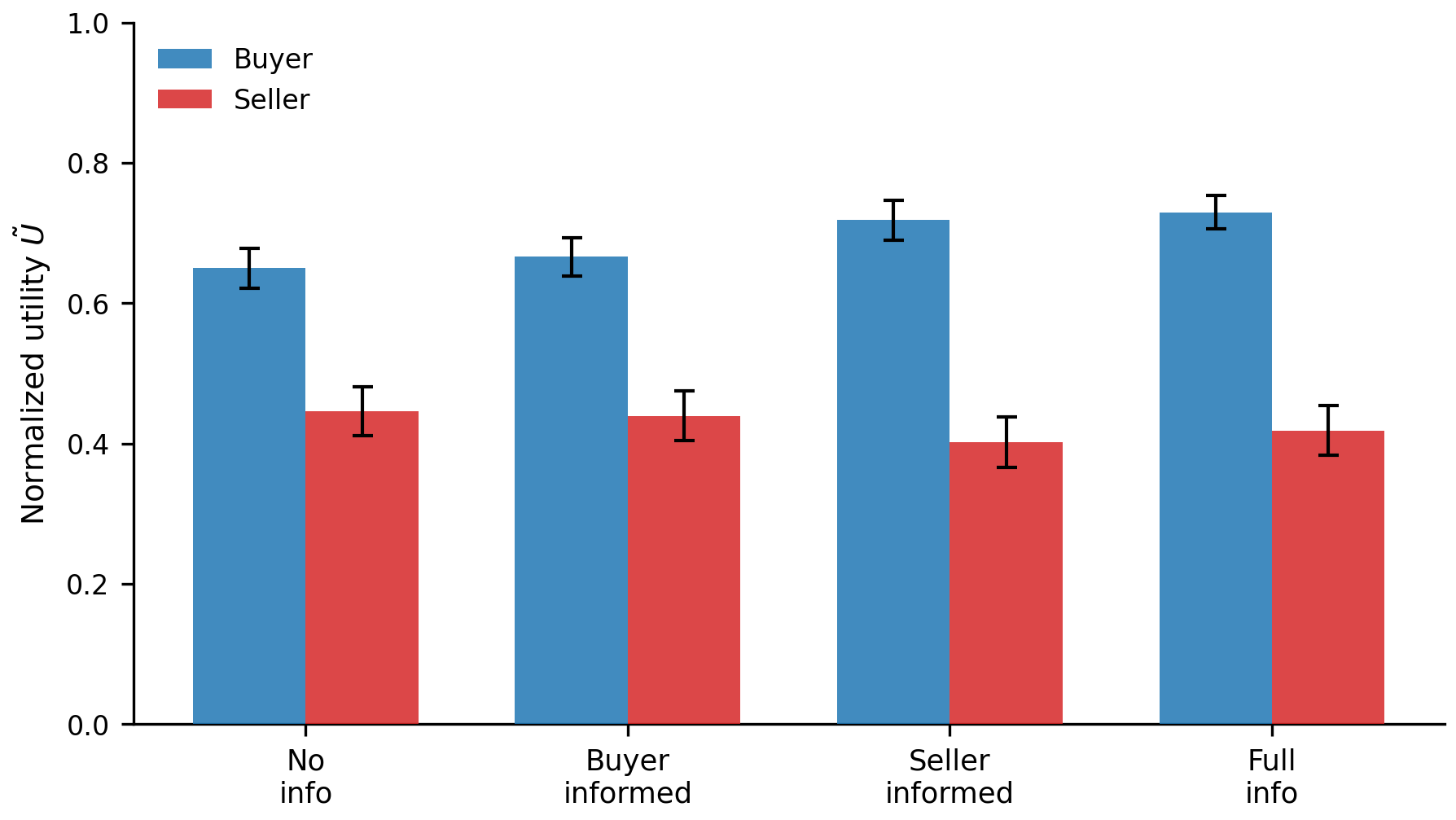}
  \caption{%
  \textbf{Buyer utility rises while seller utility declines across information conditions.}
  We show the mean normalized utility per condition, where conditions indicate whether neither side, only the buyer, only the seller, or both sides receive a ranked summary of the other party's preferences.
  Buyer utility (blue) rises monotonically with information, while seller utility (red) falls in every information condition.
  The sharpest asymmetry occurs when only the seller is informed: buyer utility increases strongly, while seller utility declines relative to the uninformed baseline.
Providing the seller with buyer's information does not improve its strategic abilities.
}
  \label{fig:outcome_asymmetry}
\end{figure}
We begin with the most conservative question: when an agent is given the partner's preferences, do final outcomes look as though that information is being used strategically? 

\Cref{fig:outcome_asymmetry} summarizes the main outcome-level pattern. We can observe that as more partner information is introduced, joint welfare increases slightly (values are shown in \Cref{tab:asym_outcomes}). But these gains do not accrue primarily to the side that receives the information. The sharpest case is the seller-informed condition: relative to the uninformed baseline, buyer utility rises while seller utility falls. Thus, information is not inert, but the resulting shifts do not match the straightforward prediction that the informed side should systematically bargain to its own advantage.

\begin{table}
  \centering
    \captionsetup{justification=centering}
\caption{%
  Outcome summary for \cond{exp\_asym}.
  $\utilde=0$ is the walk-away point and $\utilde=1$ is ideal.
  $\Delta$ is measured relative to \cond{symmetric\_none}.
   Each condition contains 100 trials in total. Values are mean (standard error).
}
  \label{tab:asym_outcomes}
  \small
  \begin{tabular}{lcrrrrrr}
    \toprule
    Condition & $\utilde_b$ (SE) & $\Delta\utilde_b$
              & $\utilde_s$ (SE) & $\Delta\utilde_s$ & Welfare & No-deal \\
    \midrule
    \cond{symmetric\_none}  & 0.650 (0.015) & ---    & 0.446 (0.018) & ---    & 1.096 & 13\% \\
    \cond{buyer\_informed}  & 0.666 (0.014) & +0.016 & 0.440 (0.018) & -0.007 & 1.106 & 12\% \\
    \cond{seller\_informed} & 0.719 (0.015) & +0.069 & 0.402 (0.019) & -0.044 & 1.120 & 22\% \\
    \cond{symmetric\_full}  & 0.729 (0.012) & +0.080 & 0.419 (0.018) & -0.028 & 1.148 & 14\% \\
    \bottomrule
  \end{tabular}
\end{table}

\begin{figure}[t]
  \centering
  \includegraphics[width=.97\linewidth]{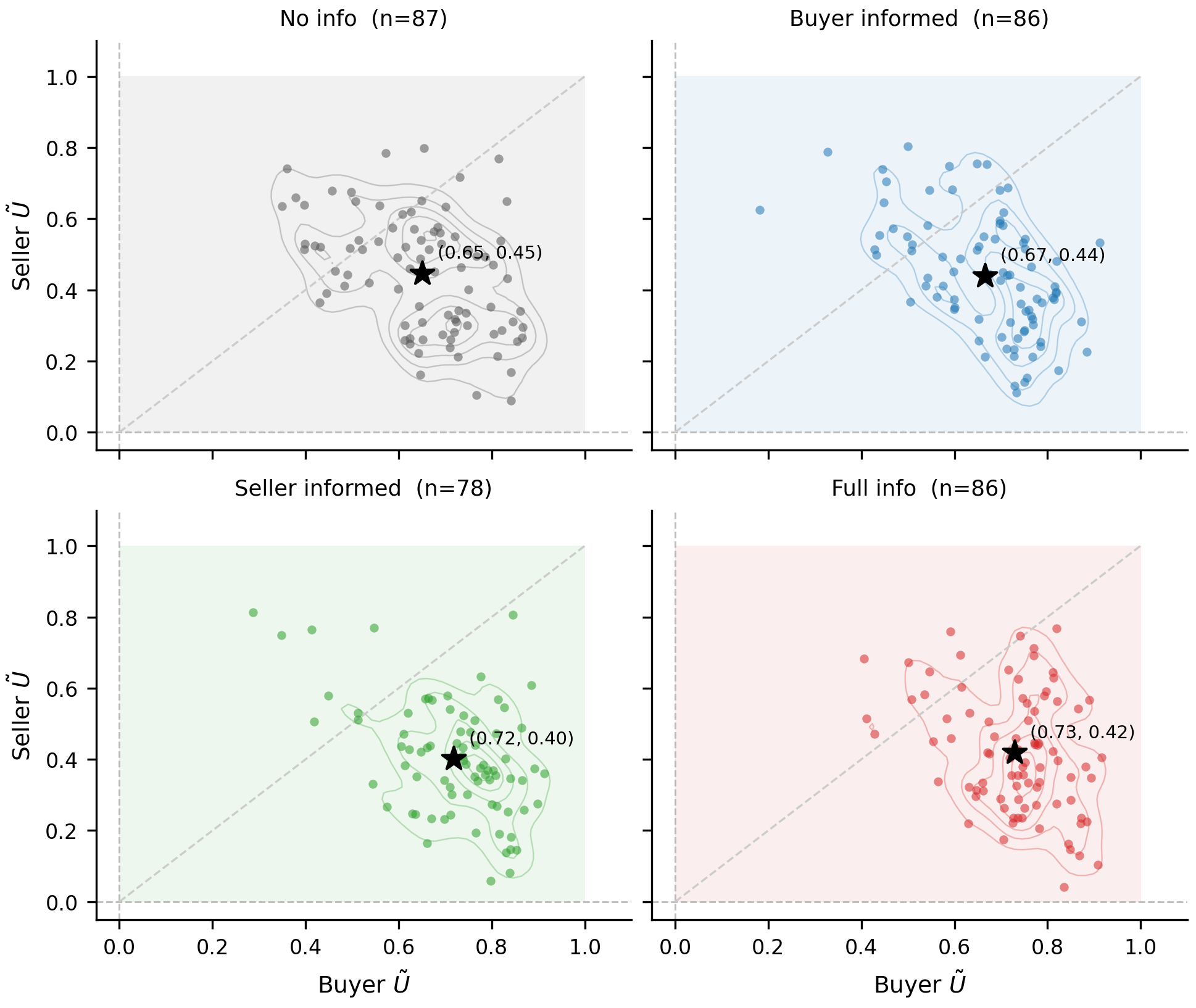}
\caption{%
  \textbf{Outcome distributions remain buyer-favorable across information conditions.}
  Each point is one agreed trial ($\bigstar$ denotes the condition mean). The x-axis is buyer normalized utility and the y-axis is seller normalized utility, so points farther right favor the buyer more and points higher favor the seller more. Across conditions, agreements remain concentrated in a buyer-favorable region, and condition means move only modestly. The clearest deviation appears in the seller-informed condition, where outcomes shift mainly toward higher buyer utility rather than higher seller utility. Counterparty information therefore changes outcomes directionally, but does not reorganize the bargaining landscape in a way that would indicate a reliable advantage for the informed side.
}
  \label{fig:quadrant}
\end{figure}

Table~\ref{tab:asym_outcomes} gives the corresponding magnitudes. Welfare increases from 1.096 in the uninformed baseline to 1.148 in the fully informed condition, but the per-side changes are small in absolute terms. The seller-informed condition is again the clearest mismatch between information access and self-benefit: buyer utility rises by $+0.069$, while seller utility falls by $-0.044$. This is far from the outcome one would expect if the informed side strategically exploited partner preference information. If an agent could use counterparty information strategically, it would choose the agreement that maximizes its own utility subject to the partner receiving at least its reservation value, thereby pushing the partner to the minimally acceptable feasible outcome. Instead, the observed gains are modest and often accrue to the uninformed side. We also observe in \Cref{fig:quadrant} that agreements remain concentrated in a buyer-favorable region across all conditions, and the condition means move only slightly as information is added. That is, information nudges the location of outcomes rather than reorganizing the bargaining landscape in a way that would clearly signal strategic exploitation.

The buyer-favorable concentration visible in \Cref{fig:quadrant} is present
even in the uninformed baseline, so information effects alone cannot account
for it. The asymmetry runs across all ten contract dimensions, not only price.
In this domain, buyer and seller utility weights are drawn independently per
trial, so for any categorical feature (vehicle model, interior finish, warranty
type, or service package) the two sides may prefer different options. When their
preferences conflict, buyers obtain their preferred categorical value in $63$ to $91\%$
of cases across these features. Sellers do not recover this ground on price: in
cases where the seller concedes their preferred model type or interior finish to
the buyer, the final agreed price is essentially the same as in cases where the
buyer is the one who conceded, differing by less than \$1k on average. Sellers
thus give up categorical features without extracting a higher price in return.
Since price accounts for only 17\% of buyer utility and the categorical features
together account for more than half, this pattern of uncompensated multi-attribute
seller concessions, not any single leverage point, explains the stable
buyer advantage observed across all conditions. 

\begin{figure}[t]
  \centering
  \includegraphics[width=0.9\linewidth]{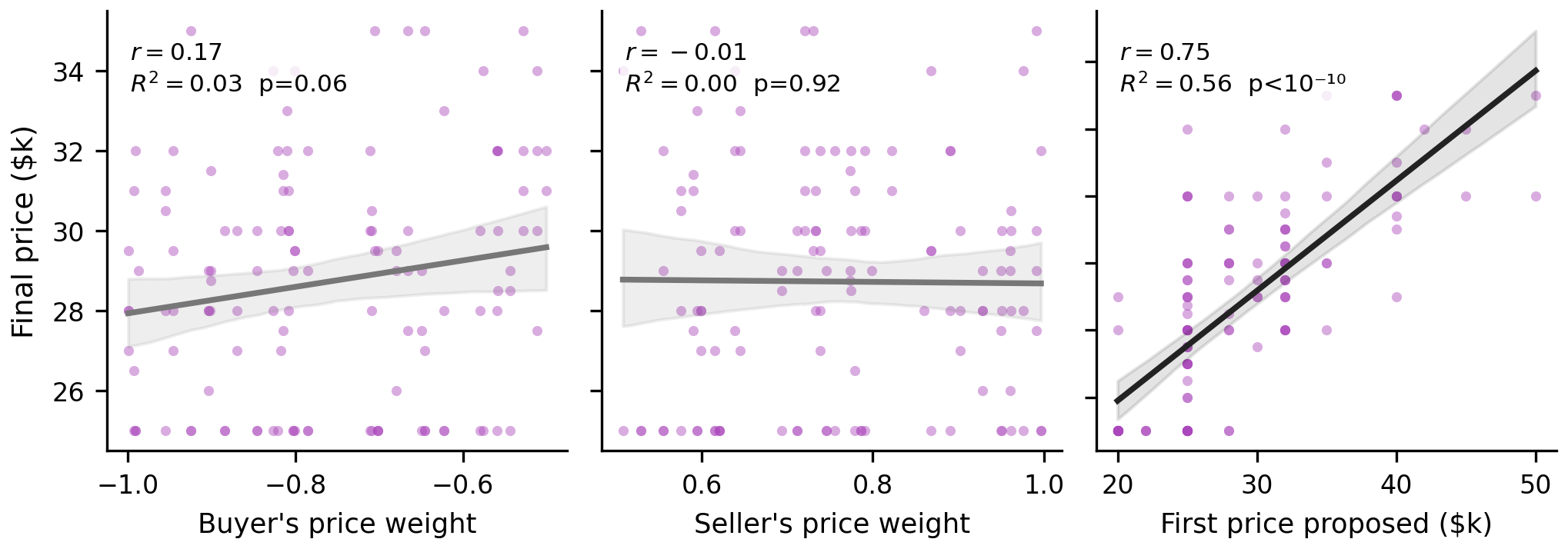}
  \caption{%
\textbf{Final price depends more on the opening price than on price utility weights.}
Trials shown ($n=123$) are pooled across all \cond{exp\_asym} conditions for cases where a Sedan was negotiated.
\emph{Left:} buyer's price weight vs.\ final agreed price.
\emph{Center:} seller's price weight vs.\ final agreed price.
\emph{Right:} first proposed price vs.\ final agreed price.
The final price is only weakly related to the buyer's and seller's price weights, whereas the opening price is a strong predictor of the final agreed price.
  }
  \label{fig:weight_vs_outcome}
\end{figure}
\FloatBarrier
In \Cref{fig:weight_vs_outcome}, we compare how the final agreed price relates to the agents' price weights versus the first price proposed during the conversation. It is clear that the final agreed price is only weakly related to the agents' underlying price utility weights, but strongly related to the first price proposed. That pattern is hard to reconcile with a process in which counterparty information is being used to construct targeted, utility-aware bargaining moves. Instead, it suggests that negotiation operates more through anchoring and local adjustment than through strategic search over reciprocal trade-offs.

\section{Counterparty Modeling is Substantial: Agents Actively Infer Preferences}
\label{sec:think_accuracy}
Given that outcome effects are too small and weakly aligned to demonstrate strategic use, we must determine where the breakdown occurs. The most basic explanation would be a failure of perception: do informed agents actually form accurate internal models of the partner's preferences? If the answer were no, then the weak strategic effects could be explained simply by failure of partner modeling. Otherwise, the problem must lie downstream, in how those beliefs are used during bargaining.

\begin{figure}[h]
  \centering
  \includegraphics[width=0.87\linewidth]{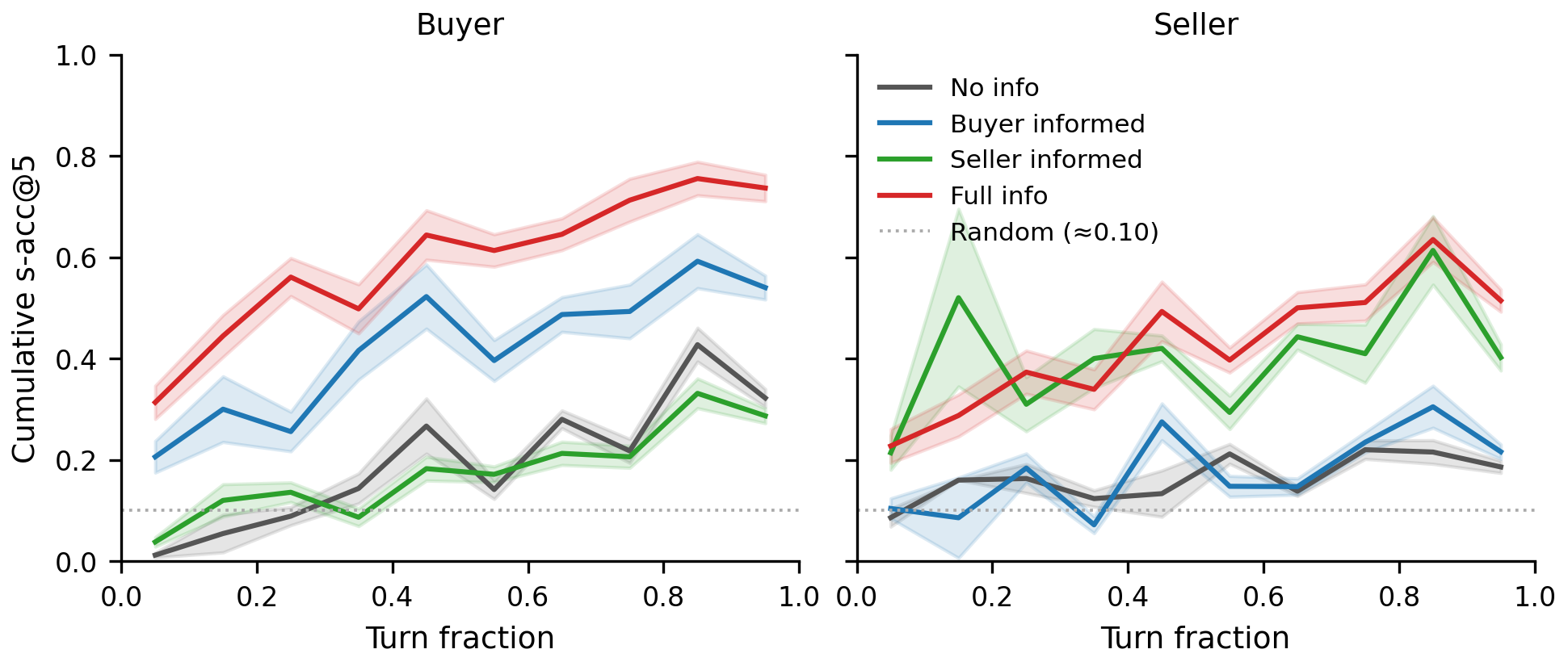}
  \caption{%
    \textbf{Informed agents rapidly form accurate negotiating partner beliefs.}
    Cumulative signed-accuracy@5 over normalized turn fraction, for buyer (left)
    and seller (right). At each turn, a prediction counts as correct when it
    identifies an attribute in the partner's true top-5 (based on the magnitude of the utility weights) with the correct preference direction, aggregated over all attributes the agent has mentioned up to that point.
    The cumulative nature of the metric means accuracy rises over turns. Note that for
    informed agents: the preference list is available from turn $1$, but agents only
    explicitly reason about individual features as they become bargaining-relevant,
    so the full partner model is articulated progressively rather than all at once.
    Informed conditions exceed $0.7$ by mid-negotiation for the buyer, while
    uninformed agents remain above chance, indicating active inference from the conversation.
    The seller-side curve rises more slowly than the buyer-side curve, suggesting weaker or less explicit incorporation of partner preferences in the reasoning trace.
  }
  \label{fig:think_accuracy}
\end{figure}

\Cref{fig:think_accuracy} shows that informed agents do, in fact, build counterparty models. The belief accuracy (described in \Cref{sec:metric}) rises quickly and remains well above the uninformed baseline for both roles, with informed conditions exceeding roughly 0.7 by the midpoint of negotiation. Even uninformed agents perform above chance, indicating that some partner inference is possible from interaction alone, but explicit information sharply improves both speed and accuracy of belief formation. The weak strategic effects in Section~\ref{sec:exp_asym} cannot be explained by saying that the models failed to notice or encode what the other side wanted. On the contrary, the traces indicate that partner information is often modeled early, explicitly, and correctly. In the example shown in Appendix~\ref{app:traces}, the informed buyer correctly identifies the seller's most important features before any offer is exchanged: higher price, Sedan, and annual service. It illustrates what \Cref{fig:think_accuracy} already shows in aggregate: the models often possess the relevant counterparty model well before the negotiation is completed.

The aggregate evidence in \Cref{fig:think_accuracy} and the concrete traces imply that the core failure is not one of perception. The agents often know what the negotiating partner values. The deeper question, then, is why these beliefs fail to organize the subsequent interaction as reciprocal, strategically coupled bargaining.

\section{Accurate Counterparty Models Fail to Translate into Strategic Execution}
\label{sec:gap}
Because informed agents do successfully model what the other side wants, the failure to secure better outcomes cannot be attributed to a lack of perception. The breakdown must therefore lie downstream in the bargaining process itself. This section asks \emph{why} this knowledge does not translate into better bargaining outcomes. Our answer is that it is a systematic process-level pattern: agents respond to inferred partner preferences, but they do not reliably convert those responses into reciprocal exchange. The resulting interaction appears to be closer to a localized accommodation behavior than bargaining.

\begin{figure}[h]
  \centering
  \includegraphics[width=0.8\linewidth]{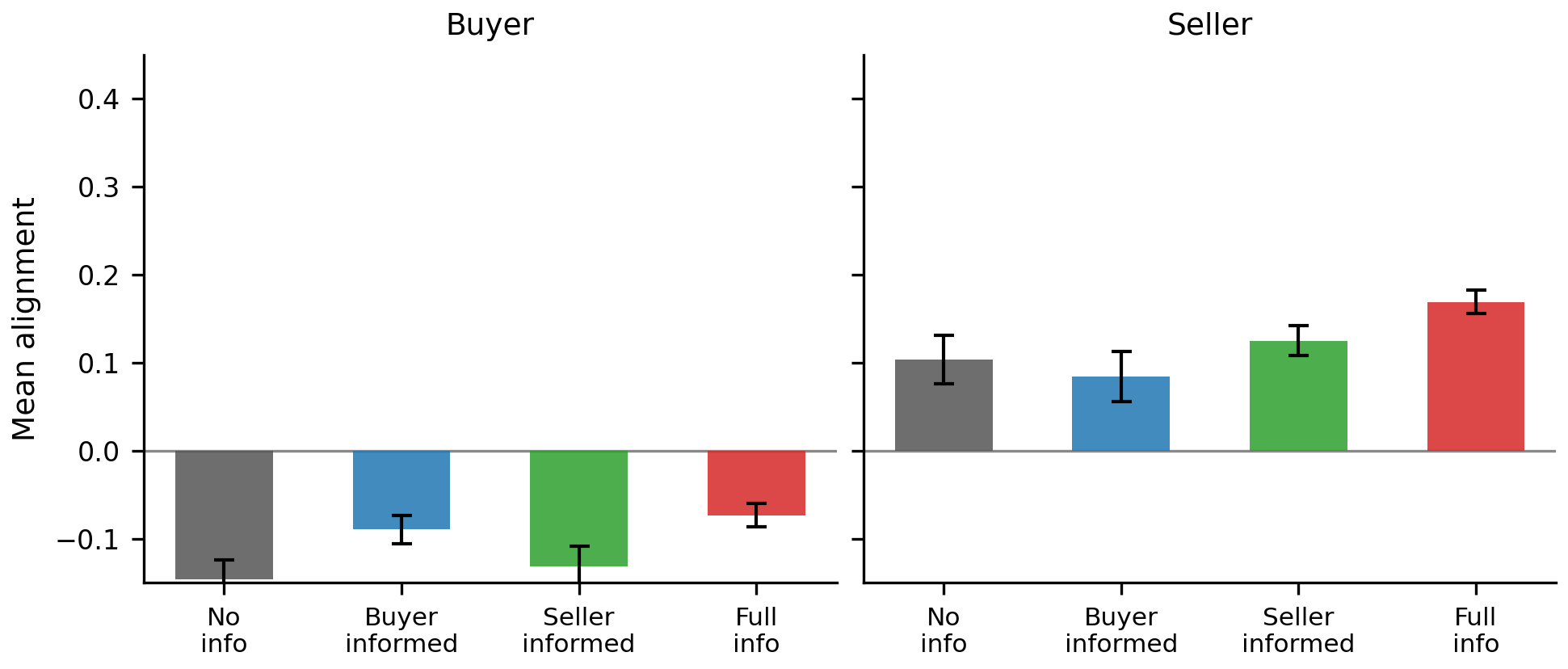}
  \caption{%
  \textbf{Sellers accommodate while buyers withhold.}
  Mean belief-action alignment by condition and role.
  Alignment is positive when an offer is closer to the direction the agent believes the partner prefers, and negative when the offer moves against the partner's inferred preference.
  Seller alignment is consistently positive, indicating accommodation of the buyer's inferred preferences, while buyer alignment is consistently negative, indicating withholding from the seller's inferred preferences.
  The asymmetry is stable across information conditions.
}
  \label{fig:alignment}
\end{figure}

We first ask whether agents move their offers in the direction they believe the partner prefers. For each turn where the \texttt{<think>} block mentions a feature preference, we compute alignment $
a = d_f \cdot (\hat{v}_f - 0.5)$, where $d_f$ is the extracted belief direction and $\hat{v}_f$ the normalized offered value. Positive alignment means accommodating the partner's inferred preference; negative alignment means moving against it. \Cref{fig:alignment} shows a clear role asymmetry. Sellers tend to move offers in the direction they believe the buyer wants, whereas buyers are more likely to move against attributes they believe the seller values. This helps explain why information does not benefit both roles symmetrically. Additional information gives the seller a more precise map of what to accommodate, while giving the buyer a more precise map of what to hold back or bargain around. This first result already shows that counterparty information is not inert. Agents do act on their beliefs about the other side. But belief-action alignment alone is not yet strategic use. A counterparty-facing concession becomes strategic only if it is paired with a compensating gain on the agent's own priorities. We therefore now ask whether turns that move the offer toward what the agent believes the counterparty wants are coupled to gains on the agent's own high-value features. If agents were using concessions strategically, moves that benefit the partner should also improve at least some of the agent's own priorities.

\begin{figure}[h]
  \centering
  \includegraphics[width=0.94\linewidth]{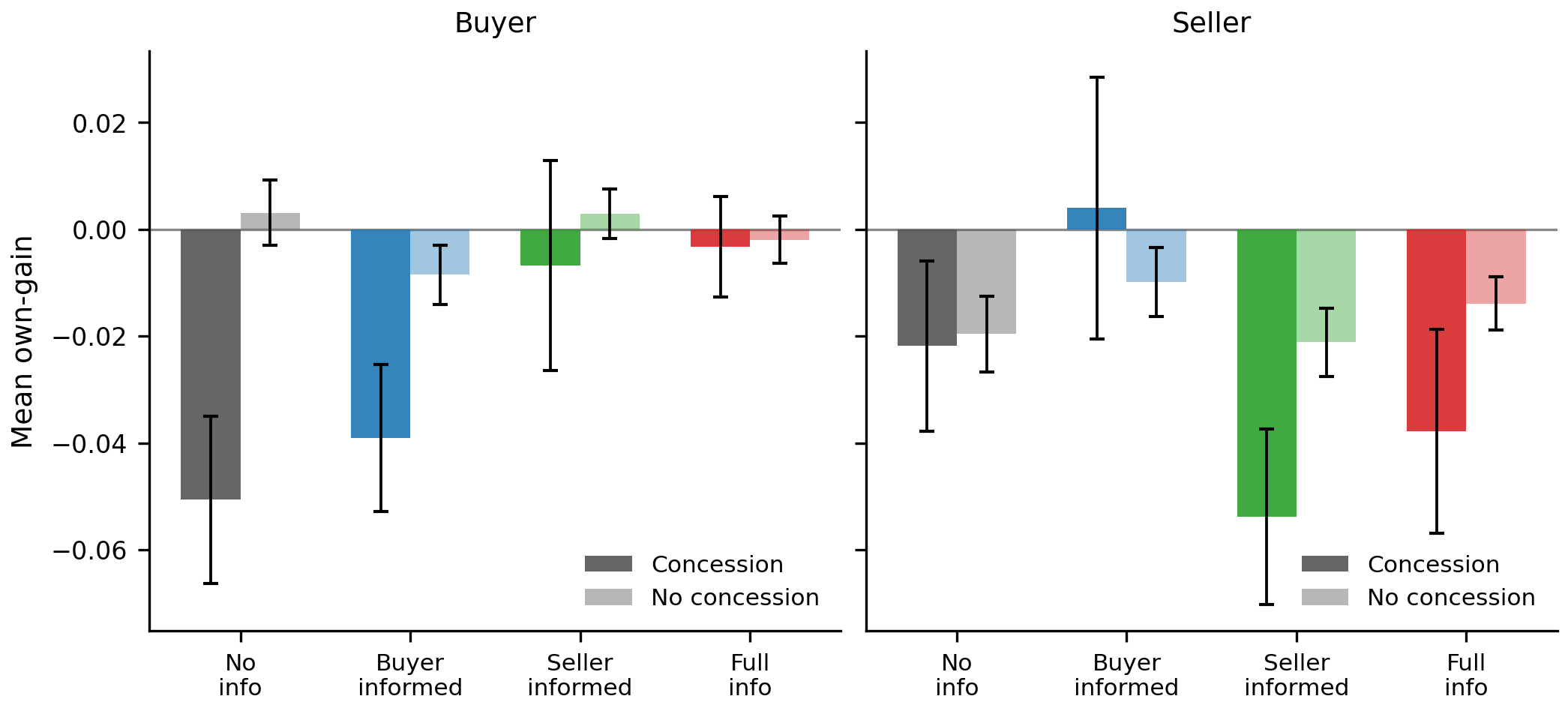}
\caption{%
  \textbf{Counterparty-facing concessions are weakly compensated by own-priority gains.}
  A counterparty-facing concession is a turn where the agent changes its offer in the direction it believes the counterparty prefers on at least one mentioned feature, so that $c_{T+1} > 0$.
  Own-gain $g_{T+1}$ measures whether the same offer revision also moves the agent's own highest-weighted features in its preferred direction.
  We compare mean own-gain on turns with a positive counterparty-facing concession against turns without one, shown separately for buyers (\textit{left}) and sellers (\textit{right}).
  If agents were using concessions strategically, concession turns should have positive own-gain.
  Instead, own-gain on concession turns is typically near zero or negative.
  Read together with \Cref{fig:alignment}, this indicates that agents often accommodate what they believe the partner wants without reliably extracting compensating gains on their own priorities.
  In asymmetric-information conditions, the informed side tends to show the weakest compensation for its concessions: the clearest drop appears for sellers in \cond{seller\_informed} and for buyers in \cond{buyer\_informed}.%
}
  \label{fig:coupling}
\end{figure}

\Cref{fig:coupling} shows that strategic coupling is weak, but not in a perfectly uniform way across roles and conditions. Own-gain on counterparty-facing concession turns is typically near zero or negative, rather than reliably positive, indicating that agents often move offers toward what they believe the partner wants without securing compensating gains on their own priorities.

Sellers are the more accommodating role overall. They more consistently move in the direction they believe the buyer prefers. Second, conditional on such concessions being made, the updated coupling analysis shows the sharper asymmetry at the level of information access. In \cond{seller\_informed}, the seller shows the clearest drop in own-gain on concession turns, while the buyer remains closer to zero. In \cond{buyer\_informed}, the same pattern appears on the buyer side: the buyer's own-gain is more negative, while the seller remains closer to zero. Thus, in asymmetric conditions, the agent with more information about the partner tends to make the more weakly compensated concessions.
The qualitative traces help unpack the aggregate mechanism. The traces in Appendix~\ref{app:traces} show that agents often identify the counterparty's priorities explicitly, sometimes revise offers in the partner's preferred direction, and can produce locally sensible trade plans that still fail to coordinate across turns. 

\Cref{fig:alignment,fig:coupling} identify the paper's central mechanism. The agents do not fail because they lack negotiating partner models. They fail because those models do not reliably structure bargaining as reciprocal exchange. What is missing is not social inference, but the ability to turn social inference into strategically coupled multi-turn negotiation.

\section{Explicit Trade Templates Do Not Close the Strategic Gap}
\label{sec:trade_plan}
If the core failure is not a lack of counterparty modeling, but rather a lack of reciprocal strategic execution, a natural intervention is to constrain the free-form interaction. We hypothesize that forcing agents to translate their beliefs into structured, explicit bargaining moves might bridge this gap. On this view, the bottleneck is not partner modeling itself, but the absence of a scaffold that forces beliefs about the counterparty to be expressed as explicit concession-for-reciprocity proposals. A structured template could plausibly help by narrowing the open-ended response space and by requiring the agent to couple each concession with a corresponding demand. This section tests that hypothesis with a templated trade-plan prompt.

In the \cond{exp\_trade\_plan} condition, agents are required to articulate an explicit exchange before each offer: a feature to concede, a feature to demand in return, and a concrete give/ask package (prompt details in Appendix~\ref{app:prompts:tradeplan}). The intervention is applied symmetrically to both agents, allowing us to ask whether making exchange structure explicit is sufficient to convert counterparty knowledge into more strategically organized bargaining.
\begin{table}[t]
  \centering
  \captionsetup{justification=centering}
  \caption{%
  Outcome metrics for \cond{exp\_trade\_plan} (100 trials per condition).
  Utilities and distances are in normalized utility space ($\utilde \in [0,1]$),
  averaged over agreed deals.
  Lower is better for $d_\text{Pareto}$ and $d_\text{NBS}$.
  The trade plan reduces deal rate for informed agents without improving
  Pareto efficiency or appreciably reducing distance to the Nash solution.
}
  \label{tab:trade_plan_results}
  \small
  \begin{tabular}{lccccc}
    \toprule
    Condition  & $\utilde_b$ (SE) & $\utilde_s$ (SE)
              & $d_\text{Pareto}{\downarrow}$ (SE) & $d_\text{NBS}{\downarrow}$ (SE) & No-deal \\
    \midrule
    \cond{uninformed\_no\_plan}  & 0.650 (0.015) & 0.450 (0.018) & 0.084 (0.007) & 0.240 (0.013) &  5\% \\
    \cond{uninformed\_with\_plan} & 0.667 (0.014) & 0.414 (0.016) & 0.095 (0.008) & 0.255 (0.013) &  5\% \\
    \cond{informed\_no\_plan}    & 0.725 (0.012) & 0.440 (0.016) & 0.053 (0.005) & 0.228 (0.015) & 13\% \\
    \cond{informed\_with\_plan}  & 0.696 (0.014) & 0.463 (0.017) & 0.058 (0.007) & 0.207 (0.014) & 20\% \\
    \bottomrule
  \end{tabular}
\end{table}

\Cref{tab:trade_plan_results} shows that the intervention does not improve the basic outcome pattern. In the informed setting, deal rate falls substantially and the utility split shifts only slightly (within one standard error) toward the seller. This is already evidence against the view that the main problem is merely one of prompt organization: forcing agents to spell out an exchange does not make negotiation more effective.

\begin{figure}[h]
  \centering
  \includegraphics[width=0.94\linewidth]{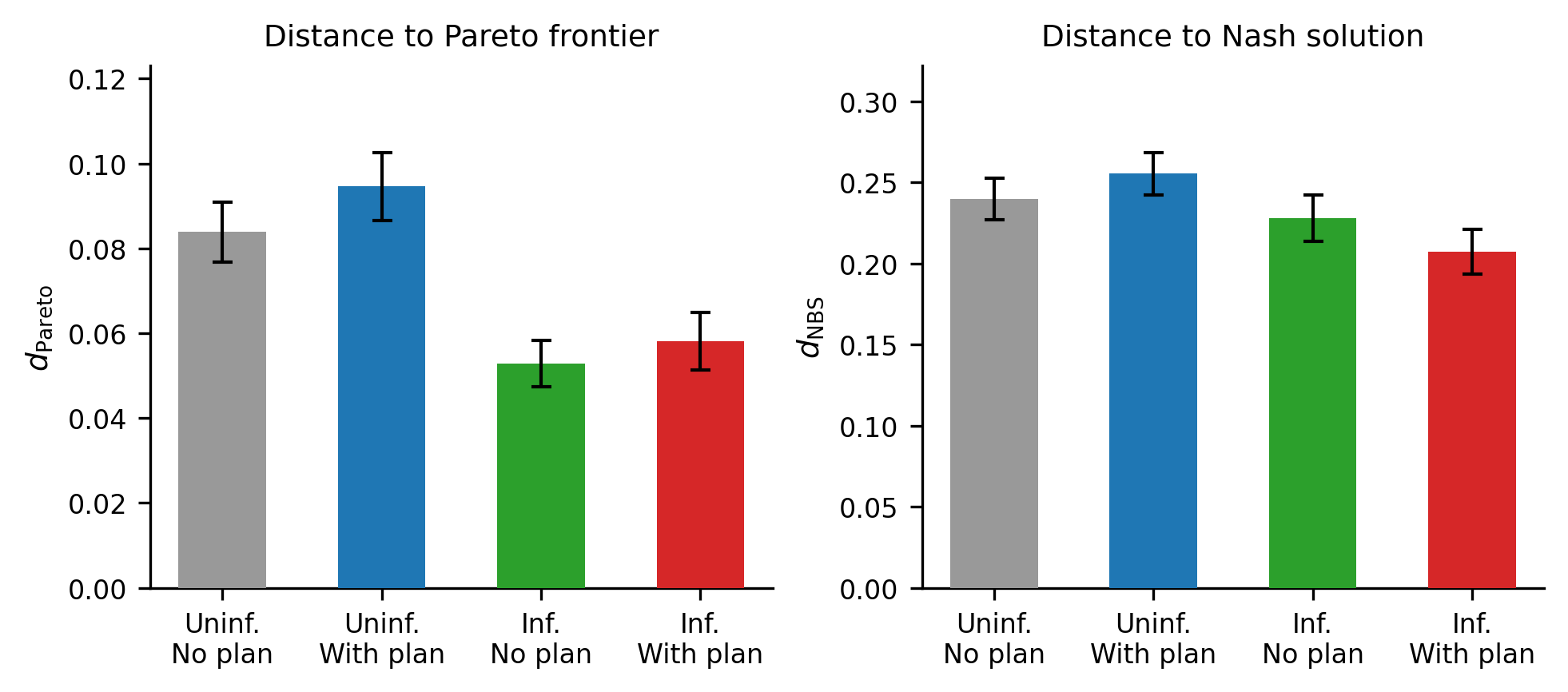}
 \caption{%
 \textbf{The trade plan does not improve efficiency.}
 The trade-plan intervention requires each agent, before making an offer, to state a feature it is willing to concede, a feature it wants in return, and a concrete give/ask package.
 \emph{Left:} distance to the Pareto frontier in normalized utility space.
 Information improves Pareto proximity ($\sim 0.09 \to 0.05$); the trade plan
 has no measurable effect on Pareto distance in either information regime.
 \emph{Right:} distance to the Nash Bargaining Solution in normalized utility
 space. The trade plan does not move informed deals appreciably closer to
 Nash (0.228 vs.\ 0.207, within one standard error) and reduces deal rate by
 about seven percentage points (cf.\ Table~\ref{tab:trade_plan_results}).
}
  \label{fig:efficiency_distances}
\end{figure}

\Cref{fig:efficiency_distances} sharpens that conclusion. If the intervention were repairing the underlying strategic failure, we would expect negotiations to move closer to efficient bargains. Instead, the trade plan has little effect on Pareto distance and no clear efficiency benefit overall. In other words, making exchange structure explicit does not cause the interaction to become more frontier-seeking or more reciprocally organized. The most plausible interpretation is that the intervention improves local neatness without improving interactive coordination. Agents can formulate explicit one-step packages, but those packages are not reliably embedded in a contingent multi-turn strategy. The problem is therefore not simply the absence of an explicit trade format; it is the failure to use such trades as part of a larger reciprocal bargaining process. The qualitative traces support the same reading. In the representative example reported in Appendix~\ref{app:traces}, the buyer proposes a locally sensible give/ask package and the seller responds with another locally sensible package, but the two packages interfere rather than reinforce. The trace matters because it makes visible what the aggregate results imply: the intervention improves legibility more than coordination.

\Cref{tab:trade_plan_results} and \Cref{fig:efficiency_distances} show that a structured trade template does not repair the gap identified in this paper. The missing capability is not the ability to state a trade, but the ability to embed such trades in reciprocal, multi-turn negotiation. This suggests that improving negotiation performance will require mechanisms that support contingent planning across turns, rather than prompt structures that only make single-turn exchanges more explicit. More generally, the result reinforces the paper’s central distinction between modeling strategically relevant information and using it to control sequential interaction.

\section{Conclusion}
\label{sec:discussion_conclusion}
In this paper, we studied whether LLM agents can use counterparty information strategically in multi-turn negotiation. Across controlled multi-attribute bargaining experiments, the main result is a gap between \emph{counterparty modeling}, a capability agents mostly possess, and \emph{strategic deployment}, a capability they lack. Agents often infer the negotiating partner's preferences accurately and early, yet that information does not reliably organize bargaining as reciprocal, contingent exchange. Instead, negotiation is shaped more by local accommodation, weak or asymmetric concession-gain coupling, and opening anchors than by multi-turn strategic coordination.

This process-level view matters because outcome metrics alone are easy to over-interpret. Providing negotiating partner information modestly shifts agreements and can slightly improve welfare, but it does not reliably benefit the side that holds the information. Once reasoning traces and turn-level offer dynamics are taken into account, a clearer picture emerges: agents respond to what the other side values, but they do not consistently convert that knowledge into bargaining leverage. Sellers are more accommodating overall, and in asymmetric-information conditions the informed side often makes the more weakly compensated concessions. The trade-plan template we introduced strengthens this interpretation. If the bottleneck were merely the absence of explicit structure, then requiring agents to formulate give/ask packages should improve bargaining performance. The template makes proposed exchanges more legible, but not more effective, suggesting that the missing capability is not simply stating a trade, but coordinating concessions contingently over multiple turns.

The broader implication is that final-outcome competence should not be mistaken for strategic competence. More generally, the presence of a capability in model reasoning or verbal reports does not guarantee the corresponding behavior in interaction. In our setting, agents can state what the counterparty wants, but that knowledge does not reliably guide reciprocal bargaining across turns. Evaluating strategic intelligence in LLM agents therefore requires testing not only what models can infer, but whether those inferences causally organize their sequential decisions. These results suggest that social understanding and strategic interaction remain importantly separable capabilities in current LLM agents. Progress will likely require methods that go beyond local response generation, including stronger models of negotiating partner response, explicit search over offer sequences, and training objectives defined over multi-turn bargaining trajectories rather than single-turn conversational quality.

\bibliographystyle{plainnat}
\bibliography{references}

\clearpage
\appendix
\small

\section{Example Traces}
\label{app:traces}

\begin{figure}[h]
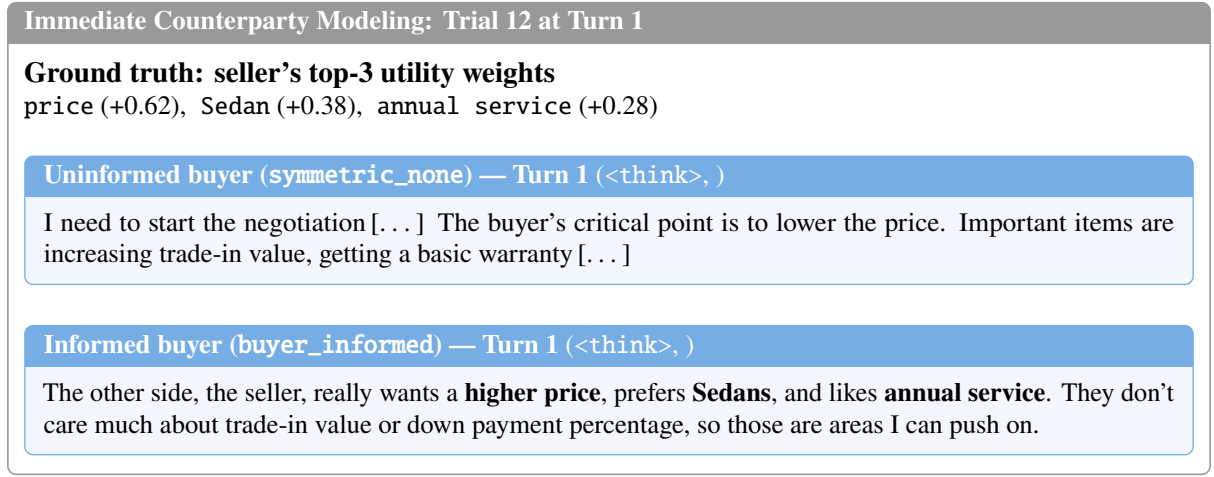

\centering
\begin{tcolorbox}[
  title={\small\textbf{Immediate Counterparty Modeling: Trial 12 at Turn~1}},
  colback=white, colframe=black!40, breakable,
  left=3pt, right=3pt, top=4pt, bottom=4pt, boxrule=0.6pt
]
\noindent\textbf{Ground truth: seller's top-3 utility weights}\\
\small
\texttt{price} (+0.62), \;\texttt{Sedan} (+0.38), \;\texttt{annual service} (+0.28)

\medskip
\begin{agentbox}[buyercolor]{Uninformed buyer (\cond{symmetric\_none}) --- Turn 1 \textnormal{(\texttt{<think>}, )}}
\small
I need to start the negotiation\,[\ldots] The buyer's critical point is to lower the price. Important items are increasing trade-in value, getting a basic warranty\,[\ldots]
\end{agentbox}
\smallskip
\begin{agentbox}[buyercolor]{Informed buyer (\cond{buyer\_informed}) --- Turn 1 \textnormal{(\texttt{<think>}, )}}
\small
The other side, the seller, really wants a \textbf{higher price}, prefers \textbf{Sedans}, and likes \textbf{annual service}. They don't care much about trade-in value or down payment percentage, so those are areas I can push on.
\end{agentbox}
\end{tcolorbox}
\caption{%
\textbf{Negotiating partner modeling.}
Before any substantive bargaining occurs, the informed buyer explicitly identifies the seller's highest-value attributes (higher price, Sedan, annual service). In fact, informed agents often possess a usable counterparty model early in the negotiation.
}
\label{fig:belief_contrast}
\end{figure}

\begin{figure}[h]
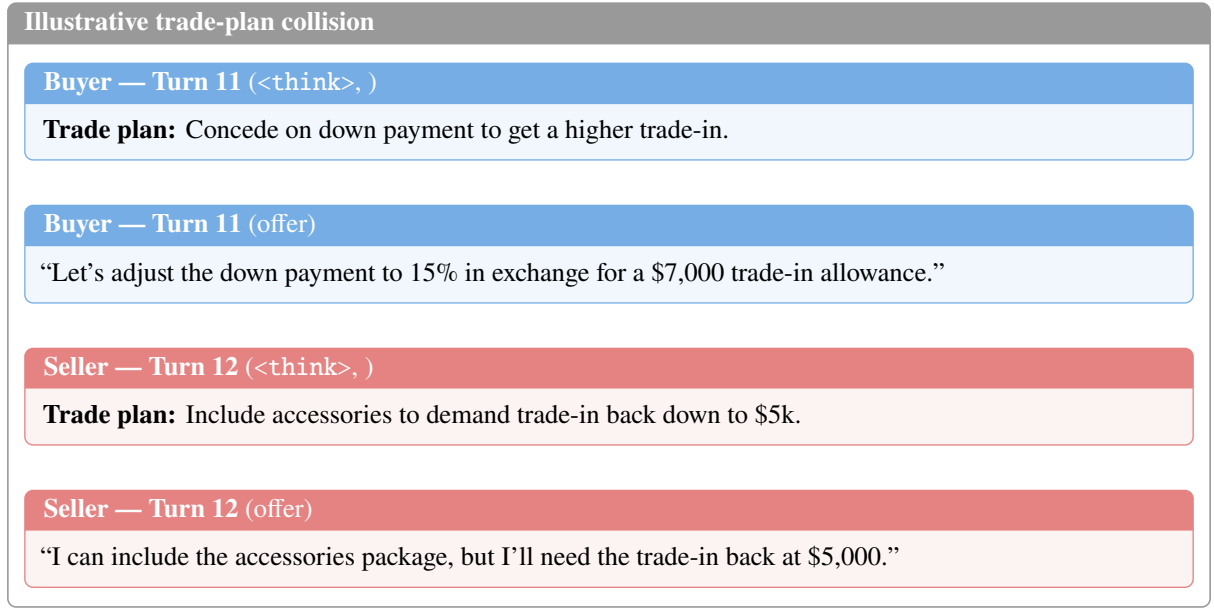

\centering
\begin{tcolorbox}[
  title={\small\textbf{Illustrative trade-plan collision}},
  colback=white, colframe=black!40,
  left=3pt, right=3pt, top=4pt, bottom=4pt, boxrule=0.6pt
]
\begin{agentbox}[buyercolor]{Buyer --- Turn 11 \textnormal{(\texttt{<think>}, )}}
\small
\textbf{Trade plan:} Concede on down payment to get a higher trade-in.
\end{agentbox}
\smallskip
\begin{agentbox}[buyercolor]{Buyer --- Turn 11 \textnormal{(offer)}}
\small
``Let's adjust the down payment to 15\% in exchange for a \$7,000 trade-in allowance.''
\end{agentbox}
\smallskip
\begin{agentbox}[sellercolor]{Seller --- Turn 12 \textnormal{(\texttt{<think>}, )}}
\small
\textbf{Trade plan:} Include accessories to demand trade-in back down to \$5k.
\end{agentbox}
\smallskip
\begin{agentbox}[sellercolor]{Seller --- Turn 12 \textnormal{(offer)}}
\small
``I can include the accessories package, but I'll need the trade-in back at \$5,000.''
\end{agentbox}
\end{tcolorbox}
\caption{%
\textbf{One-step trade plans do not compose into coordinated bargaining.}
Both agents produce locally coherent give/ask packages, but the proposed trades work at cross-purposes rather than building toward a contingent coordination.
}
\label{fig:trace_example}
\end{figure}

\section{Replication with DeepSeek-R1-671B}
\label{app:deepseek}

To assess whether the main findings depend on the particular model used in the body of the paper, we replicate the two core experiments: asymmetric information (\cond{exp\_asym}), trade-plan intervention (\cond{exp\_trade\_plan}) with DeepSeek-R1-671B~\citep{deepseek}. The negotiation domain, protocol, and evaluation pipeline are unchanged.

The main directional patterns reappear. Specifically, when the seller is informed, the buyer gains strongly and the seller loses or fails to benefit. Informed agents again achieve substantial partner-belief accuracy and the trade-plan intervention again fails to produce a clear efficiency gain. The most notable model difference is on the buyer side: with DeepSeek-R1, buyers are more accommodating than with Qwen3-235B, suggesting that the buyer-side policy is less stable across model families than the seller-side behavior.

\subsection{Asymmetric Information}
\label{app:asym_deepseek}

\begin{table}[b]
  \centering
      \captionsetup{justification=centering}
  \caption{%
    \cond{exp\_asym} outcomes for DeepSeek-R1-671B.
    Compare with Table~\ref{tab:asym_outcomes} in the main text.
    $\Delta$ is relative to \cond{symmetric\_none}.
  }
  \label{tab:asym_deepseek}
  \small
  \begin{tabular}{lcrrrrrr}
    \toprule
    Condition  & $\utilde_b$ (std) & $\Delta\utilde_b$
              & $\utilde_s$ (std) & $\Delta\utilde_s$ & Welfare & No-deal \\
    \midrule
    \cond{symmetric\_none}   & 0.640 (0.019) & ---    & 0.411 (0.027) & ---    & 1.051 & 11\% \\
    \cond{buyer\_informed}   & 0.627 (0.027) & −0.013 & 0.448 (0.030) & +0.037 & 1.075 &  5\% \\
    \cond{seller\_informed}  & 0.710 (0.015) & +0.070 & 0.395 (0.027) & −0.016 & 1.105 & 10\% \\
    \cond{symmetric\_full}   & 0.681 (0.019) & +0.041 & 0.440 (0.024) & +0.029 & 1.120 &  8\% \\
    \bottomrule
  \end{tabular}
\end{table}

 When only the seller is informed, buyer utility increases by $+0.070$, essentially identical to the $+0.069$ effect in the main Qwen experiment, while seller utility declines slightly. Welfare again rises with information. The directional pattern therefore holds: information on the seller side helps the buyer more than the seller.

The main difference appears in \cond{buyer\_informed}, where the buyer's own utility decreases slightly and the seller's utility rises. This reversal is small and within noise, but it reinforces the broader point that the information holder does not reliably capture the gains from added counterparty knowledge.

\begin{figure}[htpb]
  \centering
  \includegraphics[width=0.68\linewidth]{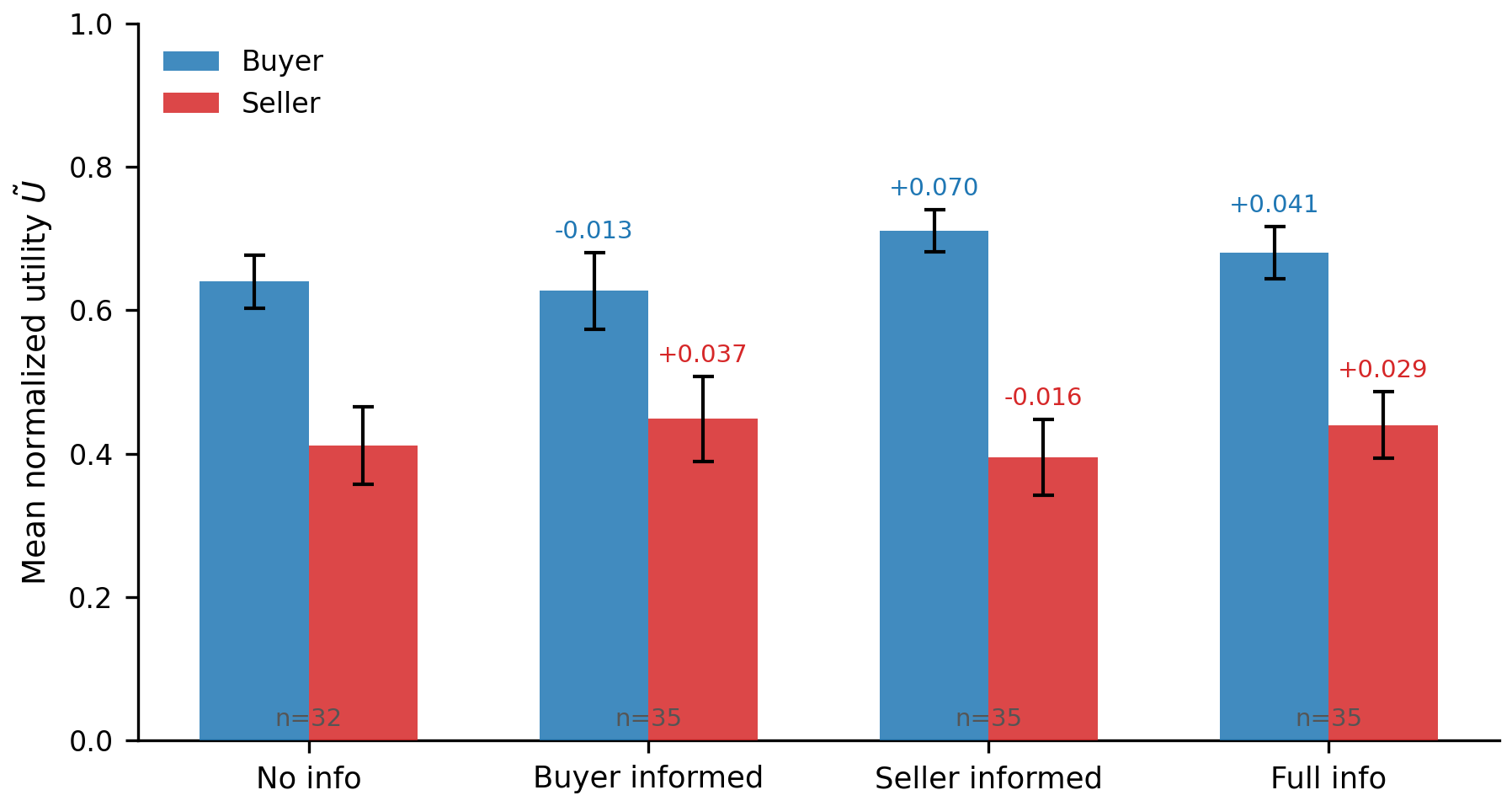}
  \caption{The seller-informed asymmetry replicates: when the seller is informed, the buyer accumulates some utility gain while the seller does not.
  }
  \label{fig:asym_deepseek}
\end{figure}

\begin{figure}[htpb]
  \centering
  \includegraphics[width=0.87\linewidth]{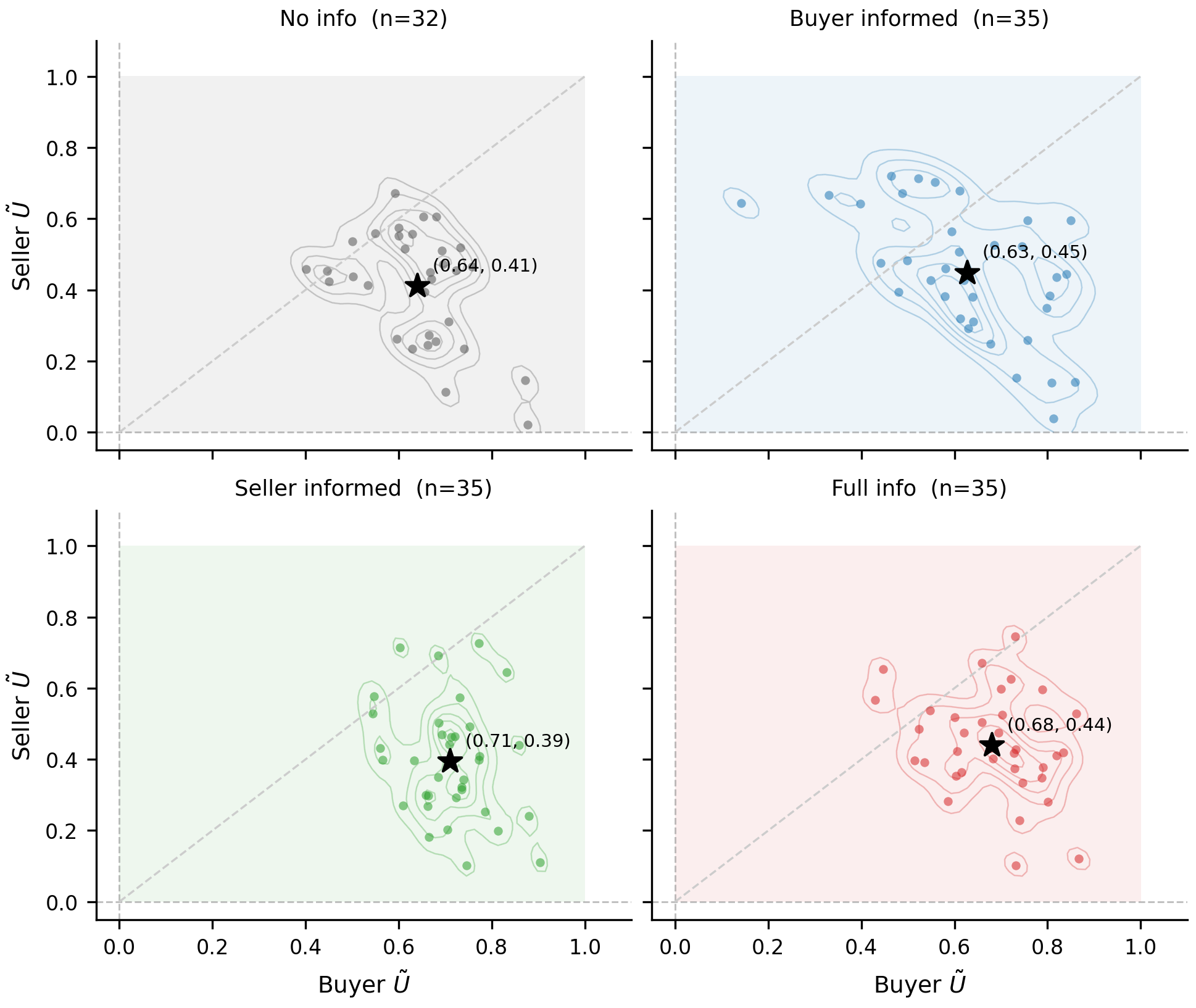}
  \caption{%
 \textbf{Final utility scatter for \cond{exp\_asym}.} Across conditions, outcomes remain concentrated in a buyer-favorable region, with the seller-informed condition shifting the mean rightward.
  }
  \label{fig:quadrant_deepseek}
\end{figure}

\subsection{Reasoning Traces: Belief Accuracy and Alignment}

\begin{figure}[tbp]
  \centering
  \includegraphics[width=0.87\linewidth]{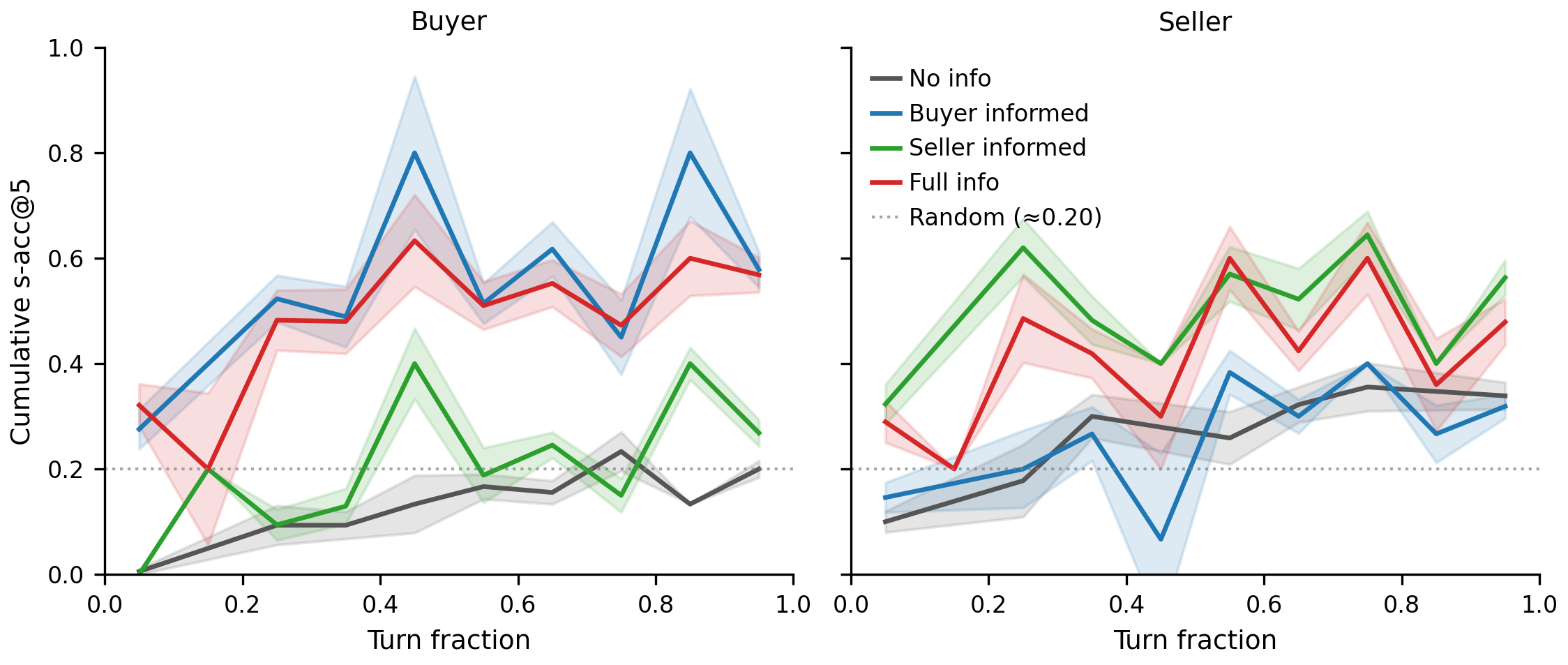}
  \caption{\textbf{ cumulative signed-accuracy@5 over normalized turn fraction.}
    Agents with access to counterparty preferences rise well above baseline, replicating the gap between accurate beliefs and strategic use.
  }
  \label{fig:think_deepseek}
\end{figure}

Figure~\ref{fig:think_deepseek} shows that agents with access to partner preferences again achieve substantial signed belief accuracy well above baseline by mid-negotiation. The basic dissociation therefore replicates: the agents can model the other side's priorities even when they do not use them to secure better outcomes for themselves.

\begin{figure}[tbp]
  \centering
  \includegraphics[width=0.94\linewidth]{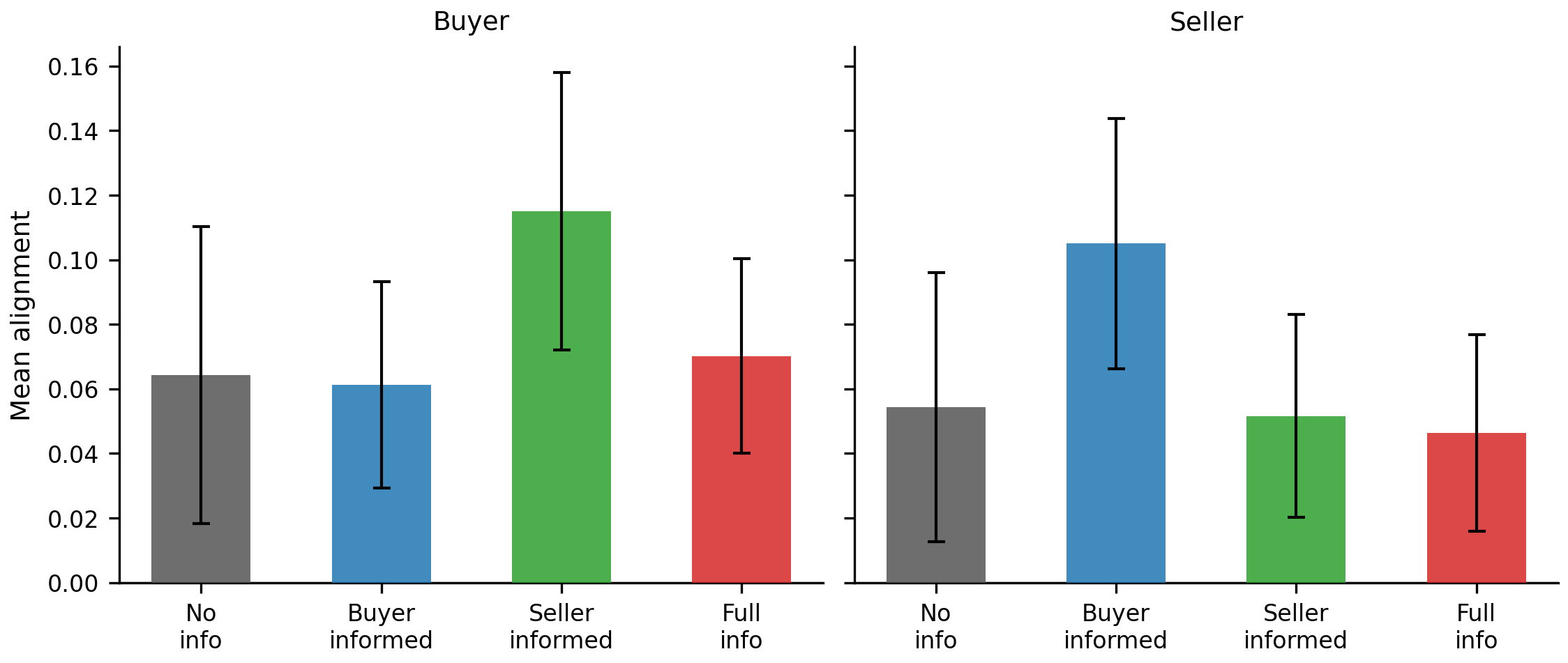}
  \caption{ \textbf{belief--action alignment by condition and role.}
    Both roles show positive alignment, indicating accommodation in both buyers and sellers.
  }
  \label{fig:alignment_deepseek}
\end{figure}

Figure~\ref{fig:alignment_deepseek} shows the main model difference. With DeepSeek-R1, both buyers and sellers have positive belief--action alignment, indicating accommodation in both roles. This differs from Qwen3-235B, where buyers tended to withhold while sellers accommodated. The robust part of the mechanism is therefore seller-side accommodation; the buyer-side response appears more model-dependent.

\begin{figure}[tbp]
  \centering
  \includegraphics[width=0.94\linewidth]{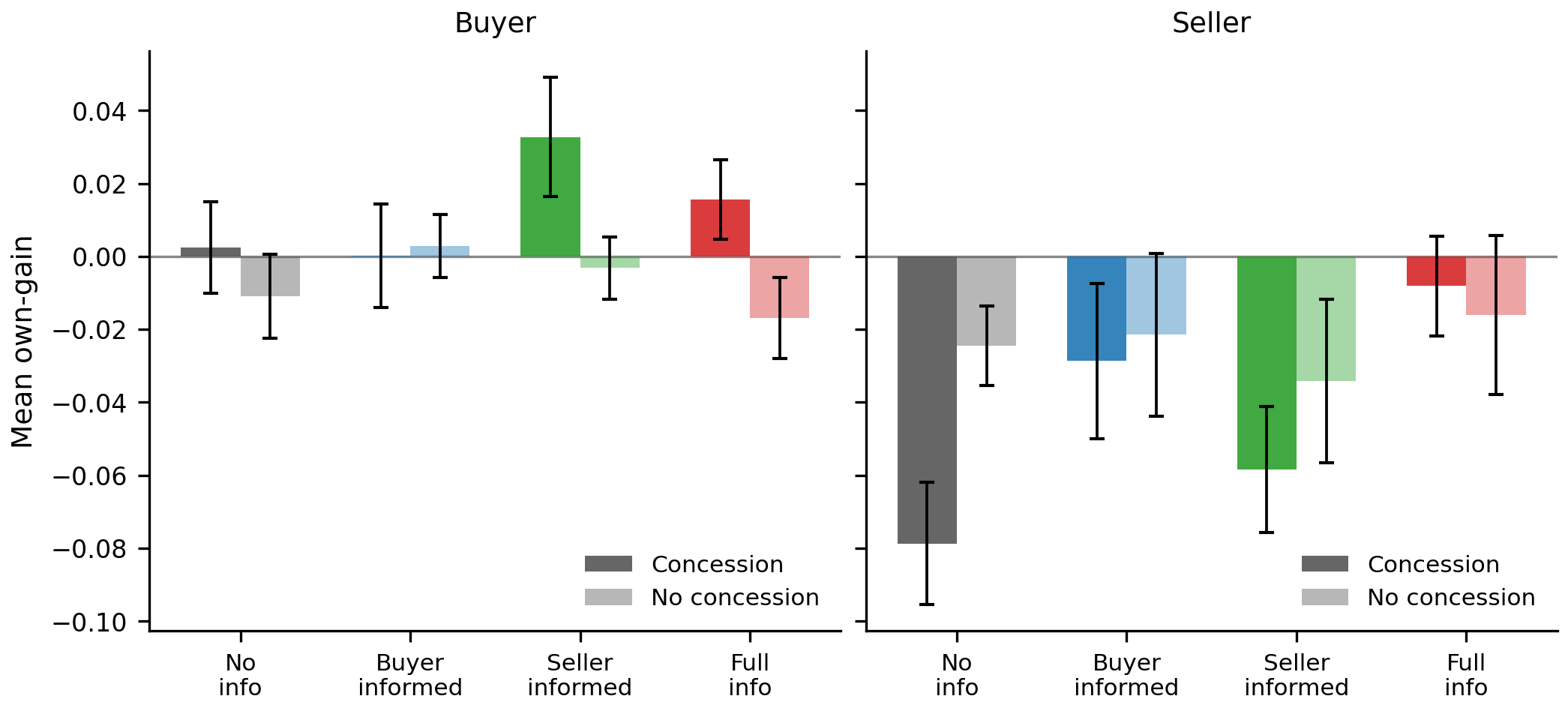}
\caption{%
  \textbf{Strategic coupling by role and condition.}
  Using concession toward counterparty $c_{T+1}$ and own-gain $g_{T+1}$ as defined in Eqs.~\ref{eq:concession}--\ref{eq:owngain}, we compare mean own-gain $g_{T+1}$ on turns with versus without positive concession toward counterparty ($c_{T+1} > 0$), shown separately for buyers (left) and sellers (right).
  The seller-side pattern replicates the main result: seller own-gain on concession turns remains negative across conditions, indicating uncompensated accommodation. The buyer side is more model-dependent: in \cond{seller\_informed}, buyers show near-zero or slightly positive own-gain on concession turns, consistent with partial strategic coupling that is absent in the main Qwen results.
}
  \label{fig:coupling_deepseek}
\end{figure}

Figure~\ref{fig:coupling_deepseek} shows a partial replication. DeepSeek sellers have negative own-gain on concession turns across all conditions, with the deepest losses in \cond{symmetric\_none} and \cond{seller\_informed}. This confirms that the core seller-side failure, conceding without extracting compensating value, is not model-specific.

The DeepSeek buyers diverge from Qwen3-235B in an informative way. In \cond{seller\_informed}, DeepSeek buyers show a small \emph{positive} own-gain on concession turns. This means that when DeepSeek buyers accommodate the seller's preferences, they can simultaneously move some of their own top-priority features in a favorable direction, a partial form of strategic coupling that Qwen3-235B buyers do not exhibit. 

\begin{figure}[htpb]
  \centering
  \includegraphics[width=0.94\linewidth]{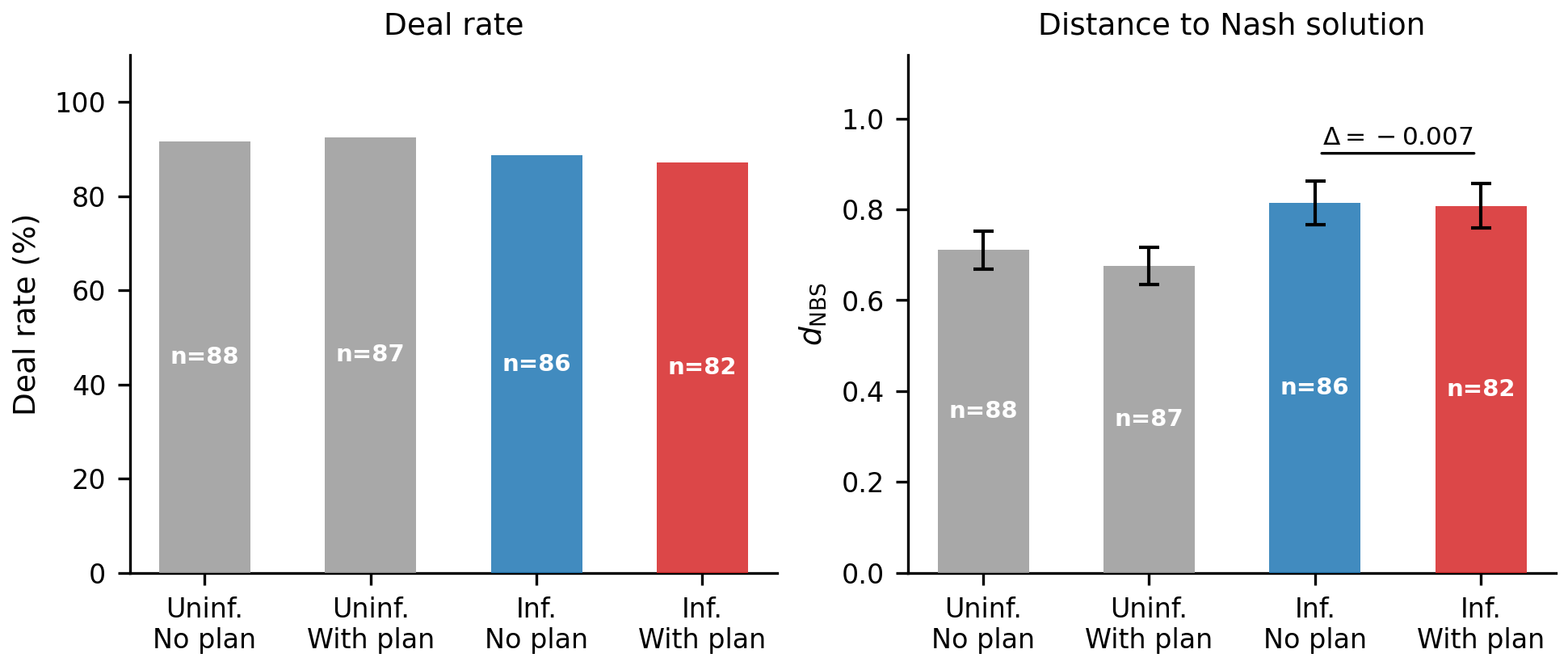}
  \caption{%
    \textbf{DeepSeek-R1 \cond{exp\_trade\_plan}: distance to Pareto frontier
    (left) and to the Nash solution (right), in normalized utility space.}
    The pattern is qualitatively the same as for Qwen3-235B: information
    improves Pareto proximity, and the trade plan has no clear efficiency
    benefit in either regime.
  }
  \label{fig:eff_deepseek}
\end{figure}

\subsection{Trade Plan Intervention}
\label{app:tp_deepseek}

\begin{table}[htpb]
  \centering
  \captionsetup{justification=centering}
  \caption{%
    Outcome metrics for \cond{exp\_trade\_plan} with DeepSeek-R1
    (100 trials per condition). Utilities and distances are in normalized
    utility space ($\utilde \in [0,1]$), averaged over agreed deals.
    Lower is better for $d_\text{Pareto}$ and $d_\text{NBS}$.
    The pattern replicates the main Qwen results: the trade plan slightly
    reduces deal rate without improving Pareto efficiency.
  }
  \label{tab:tp_deepseek}
  \small
  \begin{tabular}{lccccc}
    \toprule
    Condition & $\utilde_b$ (SE) & $\utilde_s$ (SE)
              & $d_\text{Pareto}{\downarrow}$ (SE) & $d_\text{NBS}{\downarrow}$ (SE) & No-deal \\
    \midrule
    \cond{uninformed\_no\_plan}   & 0.639 (0.013) & 0.427 (0.013) & 0.109 (0.008) & 0.233 (0.011) &  8\% \\
    \cond{uninformed\_with\_plan} & 0.609 (0.013) & 0.448 (0.017) & 0.120 (0.010) & 0.239 (0.013) &  7\% \\
    \cond{informed\_no\_plan}     & 0.671 (0.013) & 0.457 (0.016) & 0.074 (0.008) & 0.221 (0.013) & 11\% \\
    \cond{informed\_with\_plan}   & 0.663 (0.013) & 0.480 (0.018) & 0.066 (0.008) & 0.195 (0.013) & 13\% \\
    \bottomrule
  \end{tabular}
\end{table}

The trade-plan result also replicates directionally (\Cref{tab:tp_deepseek}, \Cref{fig:eff_deepseek}).
 There is no clear efficiency gain in the informed condition, and only a small change in deal rate. DeepSeek-R1 appears somewhat less brittle than Qwen3-235B under bilateral plan templates, but the main conclusion is unchanged: making one-step trades explicit does not close the gap between partner knowledge and effective bargaining.

\subsection{Qwen v.s. DeepSeek}

The most robust findings are the seller-informed asymmetry, the presence of substantial counterparty-belief accuracy, the lack of compensated seller concessions, and the failure of the trade-plan intervention to produce a clear efficiency gain. The main unstable component is buyer-side policy: Qwen3-235B buyers tend to withhold, whereas DeepSeek-R1 buyers tend to accommodate.

Across both models, the core conclusion remains stable: LLM agents can model negotiating partner preferences, but they do not reliably convert that knowledge into bargaining advantage. The seller bears the clearest cost of this gap, and structured one-step trade reasoning does not resolve it.

\section{Prompt Templates}
\label{app:prompts}


All agents use a two-level prompt architecture: $(i)$ \emph{system prompt} is set once at the start of each negotiation trial which encodes the agent's role, general strategy, domain context (available terms and valid ranges), and a trial-specific preference block, and $(ii)$ a \emph{per-turn prompt} is constructed at every negotiation step which gives the full dialogue history, the current offer on the table, and the instruction block. 

Note that, the actual prompts contain Unicode emoji markers that are not reproducible in TeX (\texttt{[!]}\ replaces a warning/critical emoji; \texttt{[>>]}\ replaces a target emoji; \texttt{[?]}\ replaces a magnifying-glass emoji; \texttt{[P]}\ replaces a lightning/plan emoji). 

\subsection{Base System Prompt}
\label{app:prompts:system}

Both buyer and seller receive the following system prompt structure. \texttt{[BUYER / SELLER]} marks role-specific text.

\begin{lstlisting}[style=promptbox]
You are a [BUYER / SELLER] negotiating to [purchase / sell] a car.

## NEGOTIATION STRATEGY:

1. **FOLLOW YOUR PREFERENCES**: You have specific preferences listed below.
   - PRIORITIZE items marked CRITICAL (most important)
   - PUSH FOR items marked IMPORTANT (but can compromise)
   - USE flexible items as bargaining chips
   - Your goal: get outcomes that match your preferences

2. **TRADE STRATEGICALLY**: Exchange things you care less about.
   - Concede on FLEXIBLE items to win on CRITICAL items
   - Don't give away things you want without getting something back
   - Propose deals that maximize YOUR outcome

3. **REACH AGREEMENT**: Making a deal is important!
   - Any deal above your reservation value is better than no deal
   - If opponent offers seem reasonable, seriously consider accepting
   - Don't let perfect be the enemy of good
   - Converge toward mutually beneficial terms

4. **UNDERSTAND CONSTRAINTS**: The other party has HARD LIMITS too!
   - They have minimum/maximum bounds they CANNOT violate
   - If they keep rejecting certain terms, you may be outside their feasible range
   - EXPLORE different combinations - don't get stuck demanding impossible terms
   - A successful deal requires finding terms that work for BOTH parties

## RESPONSE FORMAT:

Respond with: natural dialogue (2-3 sentences), then JSON. BE CONCISE.

[!] CRITICAL: Your JSON should ONLY contain terms you explicitly mentioned
in your dialogue.

For early conversation (exploring):
  {"action": "COUNTER", "terms": [], "notes": "exploring"}

For proposing specific terms you mentioned:
  {"action": "COUNTER",
   "terms": [{"name": "model", "type": "categorical", "value": "Truck"},
             {"name": "price", "type": "money", "value": 35}],
   "notes": "interested in truck around $35k"}

DO NOT include terms you haven't discussed (like color, warranty, etc.)
- let them come up naturally.

[Domain context: available car models, negotiable terms, valid ranges --
 buyer sees buyer-side ranges; seller sees seller-side minimums/ranges]

[Trial-specific preference block, see Section C.2]

NEVER mention JSON, technical details, or utility scores in your dialogue.
\end{lstlisting}

\subsection{Trial-Specific Preference Block}
\label{app:prompts:prefs}

A preference block is appended to every agent's system prompt at the start of each trial. Utility weights are drawn independently per feature from uniform ranges with sign constraints, then L1-normalized. Features are assigned to tiers by absolute weight magnitude: CRITICAL ($|w| > 0.6$), IMPORTANT ($|w| > 0.3$), FLEXIBLE (any remaining nonzero weight). Below is a representative seller example.

\begin{lstlisting}[style=promptbox]
## YOUR PREFERENCES (follow these strictly to maximize your utility):
Role: SELLER

**CRITICAL** (fight hard, don't easily concede):
  - Price: increase price (higher price -> higher utility)

**IMPORTANT** (push for):
  - Trade In: decrease trade-in value (lower value -> higher utility)
  - Delivery Day: increase delivery time (later delivery -> higher utility)

**FLEXIBLE** (use as bargaining chips):
  - Has Accessories: exclude accessories (accessories=false -> higher utility)
  - Is Truck: choose Truck (selecting Truck -> higher utility)
  - Color Blue: choose Blue color (Blue -> higher utility)
  - Interior Luxury: choose Luxury interior (Luxury -> higher utility)
  - Warranty Basic: choose basic warranty (basic -> higher utility)
  - Service Annual: choose annual service (annual -> higher utility)
  [remaining low-weight features listed similarly]

## HOW TO NEGOTIATE:
1. FOLLOW YOUR PREFERENCES - they determine your utility
2. PUSH for high-weight features (critical/important)
3. TRADE AWAY flexible items to get what you need
4. Express your preferences naturally through offers and reactions
5. Maximize utility = weighted sum of normalized features
\end{lstlisting}

\subsection{Per-Turn Prompt}
\label{app:prompts:turn}

At every negotiation step the agent receives a turn prompt assembled from the full dialogue history, the current structured offer, and a phase-dependent instruction block chosen by turn number and offer state.

\begin{lstlisting}[style=promptbox]
## CONVERSATION:
[full dialogue history]

## CURRENT OFFER ON TABLE: [last structured offer, or "None (you go first)"]

[utility info, if any]

## YOUR TURN (Turn N):
[phase-dependent instruction + optional opponent intel + optional trade plan]

Respond with: dialogue (2-3 sentences) + JSON action.

[!] REQUIRED: After any thinking, you MUST output your dialogue text and
JSON code block. Complete your full response.
\end{lstlisting}

\noindent The phase-dependent instruction evolves as follows.

\smallskip
\noindent\textit{Turn 1 (opening):}
\begin{lstlisting}[style=promptbox]
START CONVERSATIONALLY - introduce yourself and express interest.

[>>] FOR YOUR JSON:
- Just exploring? -> Use empty terms: {"action": "COUNTER", "terms": []}
- Mentioned specific things? -> Include ONLY what you said
  Example: "I'm interested in a truck" ->
    {"action": "COUNTER", "terms": [{"name": "model",
     "type": "categorical", "value": "Truck"}]}

[!] DO NOT make up values for terms you haven't mentioned yet
(like color, warranty, etc.)
\end{lstlisting}

\noindent\textit{Turns 2--5, no complete offer on table (exploratory):}
\begin{lstlisting}[style=promptbox]
DISCUSS what matters to you. Mention specific preferences.

Your JSON should include ONLY terms you explicitly mention in your dialogue.
- Example: "I'm looking at around $35k for a truck" -> Include model + price only
- Missing terms will auto-fill from their previous offer (if any)

[!] DO NOT specify terms you haven't discussed - let them emerge naturally
\end{lstlisting}

\noindent\textit{Turn 6+, no complete offer on table (propose a deal):}
\begin{lstlisting}[style=promptbox]
Time to PROPOSE A COMPLETE DEAL. State all major terms explicitly in your dialogue.

When making a complete proposal:
1. SAY all the terms in your dialogue (model, price, delivery, etc.)
2. THEN include them in your JSON
3. Don't include anything you didn't explicitly mention
\end{lstlisting}

\noindent\textit{Turns 6--15, offer on table (active bargaining):}
\begin{lstlisting}[style=promptbox]
REACT to their offer. Push for better terms or accept if good enough.

For your JSON:
- Accept their offer? -> {"action": "ACCEPT"}
- Change specific terms? -> Include ONLY the terms you want to change
- Their offer auto-fills unchanged terms

[!] IMPORTANT: The other party has HARD CONSTRAINTS they cannot violate.
If they keep rejecting certain values, try different combinations -
find the overlap zone!
\end{lstlisting}

\noindent\textit{Turns 16--30, offer on table (convergence):}
\begin{lstlisting}[style=promptbox]
CONVERGE toward a deal! Find mutually acceptable terms.

[!] If they keep rejecting your proposals, you may be outside their
feasible range.
TRY DIFFERENT TERMS - don't keep demanding impossible values.
Any deal above your reservation value is better than no deal.
\end{lstlisting}

\noindent\textit{Turn 31+, offer on table (final round):}
\begin{lstlisting}[style=promptbox]
FINAL ROUND! Accept their offer or make your final counter.

If their offer gives you positive utility (above reservation), ACCEPT IT.
Otherwise, make ONE FINAL counter and prepare to accept their response.
\end{lstlisting}

\subsection{Counterparty Intelligence Block (Informed Conditions)}
\label{app:prompts:intel}

When an agent has nonzero information quality, the following block is appended to the per-turn instruction after the phase text above. The CRITICAL / IMPORTANT / FLEXIBLE tiers mirror the preference tiers of Section~\ref{app:prompts:prefs}, but now describe the \emph{opponent's inferred preferences} with action-oriented language pointing toward the agent's own gain.

\begin{lstlisting}[style=promptbox]
## [?] INTELLIGENCE ON OPPONENT'S PREFERENCES:

**CRITICAL** (very important to them):
  - Price: decrease price (lower price -> higher utility for them)
  - Model: choose Sedan (selecting Sedan -> higher utility for them)

**IMPORTANT** (moderately important to them):
  - Interior: choose Luxury (Luxury -> higher utility for them)
  - Delivery Day: decrease delivery time (faster -> higher utility for them)

**FLEXIBLE** (they don't care much):
  - Color: choose White color (White -> higher utility for them)
  [remaining low-weight features listed similarly]

**STRATEGIC GUIDELINES - use this to MAXIMIZE YOUR OWN utility:**
- Their CRITICAL items = YOUR leverage. They need these badly ->
  demand concessions on YOUR priorities in exchange.
- Items YOU value but THEY don't care about -> Push hard here -
  giving you what you want costs them almost nothing.
- Items THEY value but YOU don't -> Only concede these in exchange for
  something YOU care about. Never give them away for free.
- Know their walk-away point: propose terms that give them JUST ENOUGH
  to accept, keeping maximum surplus for yourself.
\end{lstlisting}

\subsection{Trade-Plan Block (\texttt{with\_plan} Condition)}
\label{app:prompts:tradeplan}

In the trade-plan experiment, the following one-step reasoning scaffold is appended to every per-turn instruction of every agent (after any counterparty intelligence block). Agents are instructed to complete the template \emph{before} writing their dialogue and JSON.

\begin{lstlisting}[style=promptbox]
## [P] TRADE PLAN - complete this BEFORE writing your dialogue:

Based on what you know (opponent preferences if available, conversation
history otherwise), package a concrete trade before acting:

  STEP 1 - Feature to CONCEDE (they seem to want it, and giving it
    costs you little):
    -> Feature: ___   Direction: ___
    -> Why it's cheap for you: ___

  STEP 2 - Feature to DEMAND in return (you want it - don't give it
    away for free):
    -> Feature: ___   Direction: ___
    -> Why you should extract this now: ___

  STEP 3 - Your package: "I'll give them ___ IF they give me ___"

  (If no trade makes sense this turn, state why in one sentence.)

Only AFTER completing the plan above, write your dialogue and JSON.
\end{lstlisting}

\end{document}